\documentclass[opre,nonblindrev]{informs3}
\usepackage{mathtools, algorithm, algpseudocode}
\usepackage{enumitem}
\usepackage{bbm}
\usepackage{amsmath, amssymb}
\usepackage{amsfonts}
\usepackage{mathrsfs}
\usepackage{chngcntr}
\usepackage{bigints}
\usepackage{apptools}
\usepackage{subcaption}
\usepackage{caption}
\AtAppendix{\counterwithin{theorem}{section}}
\usepackage [english]{babel}
\usepackage [autostyle, english = american]{csquotes}
\MakeOuterQuote{"}
\usepackage{graphicx}
\usepackage{enumitem}
\usepackage{comment}
\usepackage{color}
\usepackage{float}
\usepackage{graphicx}
\usepackage{appendix}
\DoubleSpacedXI 

\usepackage{endnotes}
\let\footnote=\endnote
\let\enotesize=\normalsize
\def\notesname{Endnotes}%
\def\makeenmark{\hbox to1.275em{\theenmark.\enskip\hss}}
\def\enoteformat{\rightskip0pt\leftskip0pt\parindent=1.275em
  \leavevmode\llap{\makeenmark}}

\usepackage{natbib}
\bibpunct[, ]{(}{)}{,}{a}{}{,}%
\def\bibfont{\small}%
\TheoremsNumberedThrough     
\ECRepeatTheorems

\EquationsNumberedThrough    

\begin{document}




\TITLE{Dynamic Pricing and Demand Learning on a Large Network of Products: A PAC-Bayesian Approach}

\ARTICLEAUTHORS{%
\AUTHOR{Bora Keskin$^{1}$, David Simchi-Levi$^{2}$, Prem Talwai$^{3}$}
\AFF{${}^{1}$Duke University, Fuqua School of Business} 
\AUTHOR{\vspace{-2em}}
\AFF{${}^{2}$Institute for Data, Systems, and Society, MIT}
\AUTHOR{\vspace{-2em}}
\AFF{${}^{3}$Operations Research Center, MIT}
} 

\ABSTRACT{%
We consider a seller offering a large network of $N$ products over a time horizon of $T$ periods. The seller does not know the parameters of the products' linear demand model, and can dynamically adjust product prices to learn the demand model based on sales observations. The seller aims to minimize its pseudo-regret, i.e., the expected revenue loss relative to a clairvoyant who knows the underlying demand model. We consider a sparse set of demand relationships between products to characterize various connectivity properties of the product network. In particular, we study three different sparsity frameworks: (1) $L_0$ sparsity, which constrains the number of connections in the network, and (2) off-diagonal sparsity, which constrains the magnitude of cross-product price sensitivities, and (3) a new notion of spectral sparsity, which constrains the asymptotic decay of a similarity metric on network nodes. We propose a dynamic pricing-and-learning policy that combines the optimism-in-the-face-of-uncertainty and PAC-Bayesian approaches, and show that this policy achieves asymptotically optimal performance in terms of $N$ and $T$. We also show that in the case of spectral and off-diagonal sparsity, the seller can have a pseudo-regret linear in $N$, even when the network is dense. 
}%



\maketitle

%


\section{Introduction}

Dynamic pricing enables a seller to better manage demand over time and thereby earn higher revenues. Thanks to the widespread availability of online retail channels, a seller practicing dynamic pricing can use digital sales data to learn how demand depends on price and accordingly choose better prices. However, demand learning can be costly: choosing an exploratory price to accelerate learning the demand-price relationship results in potential revenue loss. Thus, dynamic pricing with demand learning entails a trade-off between \emph{learning} the price-sensitivity of demand to increase future revenues and \emph{earning} immediate revenues.

\paragraph{} In many industries ranging from online retail to financial services to information goods, sellers typically offer a large number of products. While this helps generate more demand volume, it also makes the demand learning problem more challenging due to increased dimensionality. This paper focuses on how the aforementioned learning-and-earning trade-off in the presence of large product networks. Specifically, we are interested in how the size of the product network moderates learning performance of a seller.

\subsection{Related Work}

There is a rich literature on dynamic pricing with demand learning. Early economics studies in this area e.g. \cite{rothschild1974two,mclennan1984price,easley1988controlling,aghion1991optimal} analyze pricing-and-learning policies in stylized settings to shed light on the optimal policy structure under simplifying modeling assumptions. The related literature in operations research and management science (OR/MS) is more recent and focuses on near-optimal or asymptotically optimal policies in more general problem settings---see, for example,  \cite{aviv2005partially,araman2009dynamic,besbes2009dynamic,farias2010dynamic,harrison2012bayesian}. This literature has been expanding in various directions, including dynamic pricing based on contextual information 
\citep{nambiar2019dynamic,chen2021statistical,ban2020personalized,golrezaei2021dynamic} 
and pricing in nonstationary demand environments 
\citep{den2015tracking,keskin2017chasing,chen2019dynamic}. 
The vast majority of the aforementioned works focus on the pricing for a single product. This paper is distinguished from these studies because we analyze dynamic pricing for a large network of products.

\paragraph{} Within the OR/MS literature discussed above, there are few studies that focus on dynamic pricing and demand learning for multiple products---see
\cite{keskin2014dynamic,den2014dynamic,keskin2019dynamic}. \cite{keskin2014dynamic} consider a linear demand model for a finite set of products and develop general sufficient conditions for asymptotically optimal learning performance. Similarly, \cite{den2014dynamic} considers pricing for a finite set of products, designing an asymptotically optimal pricing policy. \cite{keskin2019dynamic} extend to this research stream to include product design decisions: they consider a seller who designs and prices a vertically differentiated menu of products. \cite{keskin2019dynamic} construct policies that exhibit asymptotically optimal learning performance and characterize conditions under which myopic learning policies perform well in the context of vertical differentiation. Similarly, our paper considers dynamic pricing and learning for multiple products, but unlike these studies, our work sheds light on how the size and structure of the product network affects learning performance. 

\paragraph{} Our analysis is also distinct from the existing work in the broader context of online learning problems in our application of a hybrid bandit and PAC-Bayesian/Gaussian process approach to construct an optimal policy. \cite{gerchinovitz2011sparsity} studied a PAC-Bayesian approach towards sparse regression with an exponential weighting prior, which was later adapted by \cite{abbasi2011improved} for online prediction with vector parameters, however to the best of our knowledge this approach has not been applied to the matrix regression setting. Moreover, \cite{suzuki2012pac} considered Gaussian processes mixed with a PAC-Bayesian prior for i.i.d.~scalar nonlinear regression, however so far, the combination of Bayesian priors over Gaussian processes have not been considered to study sparsity in the online matrix learning setting. 

\subsection{Overview of Main Contributions}

To understand how a large product network affects the revenue performance in dynamic pricing with demand learning, we develop and analyze different structures of sparse demand-price relationships in a multi-product pricing setting. Our first contribution is the introduction of these notions to the literature on dynamic pricing with multiple products. Secondly, we design and analyze a pricing-and-learning policy that achieves asymptotically optimal performance in the sense that its pseudo-regret, i.e., revenue loss relative to a clairvoyant who knows all demand-price relationships, grows at the smallest rate in terms of the time horizon. We show that our policy's revenue loss is at most linear in the number of non-negligible demand-price relationships. Third, motivated by the pricing practice, we study a setting where the products' cross-price sensitivities are much smaller their own-price sensitivities, and find that in this setting, it is possible to achieve a tighter performance guarantee such that our policy's revenue loss is at most linear in the number of products, even when the number of non-negligible demand-price relationships is large. Fourth, we formulate and analyze a ``spectrally sparse" setting where the products' cross-price sensitivities are generated by an unknown, possibly misspecified ``similarity function", and demonstrate that under suitable smoothness conditions, our policy can again achieve a revenue loss that is at most linear in the number of products, even when the price-sensitivity matrix may be dense. 

\section{Model and Preliminaries}
We consider a firm that sells $N$ distinct products, $\{g_i\}_{i = 1}^N \subset \mathcal{G}$, over a time horizon of $T$ periods, where $\mathcal{G}$ is a metric space. In period $t \in [T] := \{1,\ldots,T\}$, the seller chooses a price $p_{t,i}$ for each product $i \in [N] := \{1,\ldots,N\}$ such that the vector of all prices in period $t$, namely $p_t = (p_{t,1},\ldots,p_{t,N})$, resides in a bounded set $\mathcal{P} \subset \mathbb{R}^N$. After choosing $p_t$, the seller observes the demand $\tilde{D}_{t,i}$ for each product $i \in [N]$ in period $t$. (For simplicity, we suppose here $\mathcal{P} = L\mathbb{B}_N$, where $L > 0$ and $\mathbb{B}_N$ is the unit Euclidean ball in $\mathbb{R}^N$). The vector of demands for all products in period $t$, namely $\tilde{D}_t = (\tilde{D}_{t,1}, \ldots, \tilde{D}_{t,N}) \in \mathbb{R}^N$, is given by
\begin{equation*}
    \tilde{D}_t = d_0 + \Theta p_t + \epsilon_t,
\end{equation*}
where $d_0 \in (0,\infty)^N \subset \mathbb{R}^N$ is the vector of baseline demands that are insensitive to prices and known to the seller, $\Theta \in \mathbb{R}^{N \times N}$ is an unknown price-sensitivity matrix, and $\epsilon_t = (\epsilon_{t,1}, \epsilon_{t,2}, \ldots, \epsilon_{t,N})$ is a vector of random demand shocks, which are unobservable to the seller. For simplicity of notation, we normalize the demand vector by letting $D_t=\tilde{D}_t-d_0$ for all $t \in [T]$. Thus,
\begin{equation}\label{eq:demand}
    D_t = \Theta p_t + \epsilon_t.
\end{equation}
For each $i, j \in [N]$, the $(i,j)^{\text{th}}$ entry of the price-sensitivity matrix, $\Theta_{i,j}$, represents the influence of product $i$'s price on the demand for product $j$. Visualizing the set of products as a network with $N$ nodes, we may interpret $\Theta$ as a ``weight matrix'' that allows self-edges. 

We assume that the demand shocks $\{\epsilon_{t,i}\}_{t \in [T], i \in [N]}$ are independent, identically distributed, and conditionally subexponential---i.e., there exist $\sigma, Q \in (0,\infty)$ such that
\begin{equation*}
    \mathbb{E}[|\epsilon_{t,i}|^n | \mathscr{F}_{t - 1}] \leq \frac{Q^{n - 2} n!\sigma^2}{2} 
\end{equation*}
for $n = 1,2,\ldots$, $t \in [T]$, and $i \in [N]$, where $\mathscr{F}_{t}:=\sigma(\epsilon_1,\ldots,\epsilon_t)$ for all $t$.

Given a price vector $p \in \mathcal{P}$, the seller's expected single-period revenue equals
\begin{equation*}
    r_{\Theta}(p) = p \cdot (d_0 + \Theta p).
\end{equation*}
For $t \in [T]$, let $H_t = (D_1, p_1, \ldots, D_{t-1}, p_{t-1}) \in \mathbb{R}^{2Nt}$ be the vectorized history of prices and observed demands until period $t$. An admissible policy $\pi$ for the seller is a sequence of functions $\{\pi_1, \pi_2, \ldots\}$ such that $\pi_t: \mathbb{R}^{2N(t - 1)} \to \mathcal{P}$ maps the history $H_{t-1} \in \mathbb{R}^{2N(t-1)}$ to a price $p_t \in \mathcal{P}$ to be charged in period $t$. Observe that the history $H_{t}$ is a random variable, and hence the seller applies the policy $\pi$ to construct a nonanticipating price sequence $\{p_1, p_2, \ldots\}$, where $p_t$ is adapted to $H_{t-1}$. The seller aims maximize its cumulative revenue, and hence, we measure the performance of a policy based on its pseudo-regret, i.e., the expected cumulative revenue loss relative to a clairvoyant who knows the value of $\Theta$ and chooses prices in accordance with this knowledge. To formally define this performance metric, let $\Theta^{*}$ be the true value of $\Theta$ and $p^{*} = \text{arg} \max_{p\in \mathcal{P}} \{r_{\Theta^{*}}(p)\}$ be the optimal price given $\Theta^{*}$. Then, the seller's pseudo-regret is
\begin{equation*}
    \Delta_T = \sum_{t = 1}^T r_{\Theta^{*}}(p^{*}) - r_{\Theta^{*}}(p_t),
\end{equation*}
where $p_t$ is the price charged in period $t$. The seller's goal is minimize $\Delta_T$ while learning $\Theta^{*}$. We study this problem under different sparsity assumptions on the underlying product network. Specifically, we consider three notions of sparsity: $L_0$ sparsity, off-diagonal sparsity, and spectral sparsity. 

\subsection{$L_0$ sparsity}
In our first framework, we suppose that there is sparse set of demand-price relationships in the product network. That is, the unknown price-sensitivity matrix $\Theta$ is $L_0$ sparse in the sense that the number of nonzero entries of $\Theta$ is substantially less than the number of its total entries, $N^2$. Here, we design a pricing policy achieving a $\mathcal{O}(\sqrt{sNT}$) regret rate where $s$ is the number of nonzero entries in $\Theta$. 

\subsection{Off-Diagonal Sparsity}
According to the demand model in \eqref{eq:demand}, the diagonal entries of $\Theta$ represent the products' own-price sensitivities, whereas the off-diagonal entries of $\Theta$ represent cross-price sensitivities. In practice, the demand for a product is typically more sensitive to the product's own price than to other products' prices. Motivated by this, we consider a \textit{specified} sparsity setting, where we suppose that the diagonal entries are $\mathcal{O}(1)$ and the off-diagonal entries are $\mathcal{O}(1/N)$ as $N$ grows large. We demonstrate that in this setting, we are able to design a policy achieving the asymptotically optimal $\mathcal{O}(N\sqrt{T})$ regret rate even when $\Theta$ may be full. 

\subsection{Spectral Sparsity}
In our second framework, we incorporate the geometry of the product space $\mathcal{G}$. At a high level, we suppose the sensitivity of the demand of product $x$ on the price of product $y$ reflects the topology of $\mathcal{G}$, i.e. products that are close in $\mathcal{G}$ are highly sensitive to each others' demands. In practice, this can correspond to forming product categories based on product specifications and features, and using this information on estimating the price-sensitivity of demand.

\paragraph{} Specifically, we suppose that $\Theta$ is generated by some (possibly anisotropic) function $\kappa^{*}: \mathcal{G} \to \mathbb{R}$, i.e. for products $g_i, g_j \in \Omega \subset \mathcal{G}$, the entry $\Theta_{ij}$ in the price-sensitivity matrix $\Theta$ is given by $\Theta_{ij} = \kappa^{*}(g_i - g_j)$. In related literature, this function is often referred to as a kernel, however this typically requires imposing stronger conditions on $\kappa^{*}$ (such as positive definiteness) which we avoid here. Instead, we simply think of $\kappa^{*}$ as some unknown generating function we wish to learn. Moreover, we require that the subset $\Omega$ containing our $N$ products is bounded and sufficiently regular (the precise notion of regularity required will often depend on the problem setting). 

\paragraph{} Here, we wish to learn $\kappa^{*}$ under a \textit{misspecified} framework, where the seller does not know a precise hypothesis space containing $\kappa^{*}$. Instead, given a subspace $\mathcal{H} \subset L^2(\Omega)$, the seller knows $\kappa^{*}$ lies in some \textit{interpolation space} lying between $\mathcal{H}$ and $L^2(\Omega)$. Specifically, we suppose:
\begin{equation*}
    \kappa^{*} \in [L^2(\Omega), \mathcal{H}]_{\alpha, 2}
\end{equation*}
for some unknown $\alpha \in (0, 1)$ (see e.g. \cite{adams2003sobolev} for a definition).

\begin{remark} 
Note when $\alpha = 1$, we revert back to the \textit{correctly specified} situation and $[L^2(\Omega), \mathcal{H}]_{1, 2} = \mathcal{H}$, i.e., the hypothesis space $\mathcal{H}$ indeed realizes the learning target $\kappa^{*}$. 
\end{remark}

\paragraph{} In our analysis, we suppose $\mathcal{H}$ is a separable reproducing kernel Hilbert space (RKHS) compactly embedded into $L^2(\Omega)$, with bounded kernel $\mathcal{K}$ (we denote this $\mathcal{H}_{\mathcal{K}}$ or $\mathcal{H}$ when the kernel is understood). The choice to consider $\mathcal{H}$ as an RKHS stems from the intimate connection between RKHS interpolation spaces and the sample paths of Gaussian processes (see \cite{kanagawa2018gaussian} and \cite{steinwart2019convergence}), which will be crucial in the subsequent analysis of our PAC-Bayesian estimator. We briefly summarize this relationship here. Let $I_{\mathcal{K}}: \mathcal{H}_{\mathcal{K}} \to L^2(\Omega)$ denote the compact embedding of the RKHS into $L^2(\Omega)$ and $I_{\mathcal{K}}^{*}: L^2(\Omega) \to \mathcal{H}_{\mathcal{K}}$ its adjoint. By the spectral  theorem for compact, self-adjoint operators we have that the operator $T_{\mathcal{K}} = I_{\mathcal{K}}I^{*}_{\mathcal{K}}$ on $L^2(\Omega)$ enjoys the Mercer representation:
\begin{equation*}
\label{Embedding Operator}
    T_{\mathcal{K}} = \sum_{i = 1}^{\infty} \mu_i e_i \otimes e_i,
\end{equation*}
where $\{\mu_i\}_{i = 1}^{\infty}$ are the eigenvalues of $T_{\mathcal{K}}$ and $\{e_i\}_{i = 1}^{\infty} \subset L^2(\Omega)$ an orthonormal basis of eigenvectors (note, formally $\{e_i\}_{i = 1}^{\infty}$ is a family of equivalence classes in $L^2(\Omega)$ --- for notational simplicity, we omit this formalism here). Equipped with the spectral representation of $T_{\mathcal{K}}$, we may define the RKHS power spaces $H_{\mathcal{K}}^{\beta}$:
\begin{equation}
\label{Interpolation Definition}
    \mathcal{H}_{\mathcal{K}}^{\beta} = \left\{f = \sum_{i = 1}^{\infty} a_i \mu^{\frac{\beta}{2}}_i e_i: \{a_i\} \in \ell^2 \right\}
\end{equation}
\paragraph{} It can be easily seen that the power spaces $\mathcal{H}_{\mathcal{K}}^{\beta}$ are descending with $\mathcal{H}_{\mathcal{K}}^{\beta_1} \subset \mathcal{H}_{\mathcal{K}}^{\beta_2}$ for $\beta_1 > \beta_2$, $\mathcal{H}_{\mathcal{K}}^1 = \mathcal{H}$, and $\mathcal{H}_{\mathcal{K}}^0 = L^2(\Omega)$ (the latter two statements hold as we assume $\{e_i\}_{i = 1}^{\infty}$ to be an orthonormal basis). We denote by $\mathcal{K}^{\beta}$ the kernel of the integral operator $T^{\beta}_{\mathcal{K}}$. Note, that although $\mathcal{H}_{\mathcal{K}}^{\beta}$ is a Hilbert space, it is not necessarily a RKHS, which additionally requires $\sum_{i = 0}^{\infty} a_i \mu^{\beta}_i e^2_i(x)$ to be almost-surely finite (see \cite{steinwart2019convergence} for details). It is well-known that $\mathcal{H}_{\mathcal{K}}^{\beta}$ is equivalent to $[L^2(\Omega), \mathcal{H}_{\mathcal{K}}]_{\beta, 2}$ (\cite{steinwart2019convergence, steinwart2012mercer}), which provides a convenient, intuitive characterization of the latter abstract interpolation space. This powerful characterization has been harnessed to study the sample paths of Gaussian processes, whose smoothness properties can be related to the interpolation spaces $\mathcal{H}_K^{\beta}$ via the Karheune-Loeve expansion of the process. A comprehensive treatment of these connections may be found in \cite{kanagawa2018gaussian} and \cite{steinwart2019convergence}. 

\paragraph{} In our analysis, we impose an exponential decay condition on the eigenvalues: $\mu_i \asymp e^{-qi}$ (where $q > 0$). This decay rate is quite commonly considered/observed in the literature on regularized learning (see e.g. \cite{chatterji2019online} and \cite{zhou2002covering}), and satisfied by several common $C^{\infty}$ kernels such as the square-exponential and inverse multiquadric kernels on sufficiently regular domains $\Omega$ (see \cite{belkin2018approximation} for a detailed discussion). We note that while the eigenvalues $\{\mu_i\}_{i = 1}^{\infty}$ are not always known explicitly, their asymptotic tail behavior is often well-understood. Moreover, in certain cases, the interpolation spaces $\mathcal{H}^{\beta}$ themselves are RKHSs, whose reproducing kernels $\mathcal{K}^{\beta}$ are readily available. A notable example is the isotropic square-exponential (Gaussian) kernel of bandwidth $\sigma > 0$ on $\Omega \subset \mathbb{R}^d$ (see \cite{steinwart2008support} for a definition) whose Mercer expansion in \eqref{Embedding Operator} has a well-known closed form, which enables the straightforward computation of the interpolation spaces in \eqref{Interpolation Definition}. Intuitively, the exponent $\beta$ in \eqref{Interpolation Definition} characterizes the ``smoothness'' of the functions lying in $\mathcal{H}^{\beta}$ (with larger values of $\beta$ indicating greater smoothness). Roughly, a Gaussian process with Cameron-Martin space $\mathcal{H}$, will possess sample paths that are less smooth than the functions in $\mathcal{H}$, i.e., they will reside in the interpolation space $\mathcal{H}^{\beta}$ for some $\beta \in (0, 1)$ (see Theorem 5.2 in \cite{steinwart2019convergence} for a precise statement).  

\paragraph{} We now explain the basis for choosing the term ``spectral sparsity" to describe this setting. From the definition of $\mathcal{H}_{K}^{\alpha}$, our assumption that $\kappa^{*} = \sum_{i = 1}^{\infty} a_i e_i \in \mathcal{H}^{\alpha}$ translates into the requirement $\sum_{i=1}^{\infty} \frac{a_i}{\mu_i^{\alpha}} < \infty$. Since $\mu_i \asymp e^{-iq}$, by assumption, we obtain or $a_i = o(e^{-i \alpha q})$, i.e., the ``Fourier coefficients'' (spectra) of $\kappa^{*}$ with respect to the orthonormal basis $\{e_i\}_{i=0}^{\infty} \subset L^2(\Omega)$ decays at least exponentially, i.e. they exhibit ``sparsity''. In the classical example where $\Omega$ is isomorphic to $[-\pi, \pi]^d$ for some $d > 0$, this notion corresponds precisely to the exponential decay of the Fourier coefficients and hence the members of $\mathcal{H}$ are infinitely smooth (and characteristically exhibit rapid decay in the isotropic/one-dimensional case, see \cite{minh2010some}). It is important to note that sparsity here does \textit{not} refer to the number of nonzero coefficients, but merely their asymptotic behavior. Indeed, under the spectral sparsity framework, we allow $\Theta^{*}$ to potentially be a full matrix, and observe that our learning rates do not depend on the number of nonzero entries in $\Theta$, but merely the smoothness properties of the generating function $\kappa^{*}$. We are able to design a policy achieving the asymptotically optimal $\mathcal{O}(N\sqrt{T})$ regret rate even when $\Theta$ may be full. The term ``spectra" is taken in analogy to the kernel learning literature, in which $\kappa^{*}$ is further assumed to be positive-definite and isotropic, and the aforementioned Fourier coefficients are called the ``spectral density'' of $\kappa^{*}$. 

\paragraph{} The smoothness condition on the generating function $\kappa^{*}$ stems from the assumption that the price sensitivities (edge weights) in $\Theta_{ij}$ reflect the underlying geometry of the feature space $\mathcal{G}$. The latter assumption may be quantified by requiring $\sum_{i, j = 1}^N |\Theta_{ij}|||g_{i} - g_{j}||^2_{\mathcal{G}}$ to be small. Indeed, this implies that $||g_{i} - g_{j}||^2_{\mathcal{G}}$ is small when $|\Theta_{ij}|$ is large, i.e. when products $i$ and $j$ are strongly connected they are close in the product space. This notion has been harnessed recently for the problem of graph learning: given data $\{g_i\}_{i = 1}^N$ (i.e. product features), \cite{kalofolias2016learn} suggests learning the graph $\Theta$ by minimizing the following objective:
\begin{equation}
\label{Graph Learning}
   \Theta = \text{arg} \min_{\bar{\Theta}} \sum_{i, j = 1}^N \bar{\Theta}_{ij}||g_{i} - g_{j}||^2_{\mathcal{G}} + h(\bar{\Theta}_{ij})
\end{equation}
for some regularizer $h$ (note in \eqref{Graph Learning}, $\bar{\Theta}_{ij}$ is assumed nonnegative, which is relaxed in our problem setting). When $h$ is chosen as $h(x) = \sigma^2x(\log x - 1)$, \cite{kalofolias2016learn} demonstrates that we recover the isotropic $C^{\infty}$ generating function $\Theta_{ij} = \text{exp}\Big(\frac{||g_{i} - g_{j}||^2_{\mathcal{G}}}{\sigma^2}\Big)$. Moreover, compared to non-smooth regularizers like the $L_1$ norm, the function $h$ used here is undefined for nonpositive arguments (i.e. $W_{ij}  \neq 0$), making the spectral sparsity framework highly suitable for considering frameworks where the connections between products are ubiquitous but decaying in intensity with distance in $\mathcal{G}$. In a sense, our framework can be viewed as a generalization of the above graph learning approach, in which we do not specify a regularizer $h$ in the constraint \eqref{Graph Learning}, and desire to simultaneously learn a $\Theta$ that both reflects our demand model and the geometry of the product space. 

\begin{remark}
We note that assuming the existence of a generating function $\Theta_{ij} = \kappa^{*}(g_i - g_j)$ implies that $\Theta$ is constant on the diagonal. This restriction may be easily overcome in practice by learning the diagonal entries of $\Theta$ prior to learning the full matrix. The latter task can be accomplished by performing $N$ independent one-dimensional regression tasks --- i.e. to learn $\Theta_{ii}$ we repeatedly set $p_{t} = e_i$ ($i^{th}$ standard basis vector) and observe $D_{t, i}$. Once the diagonal elements have been learned (with a regret of order $N \log T$), we may learn the off-diagonal entries in the second phase by shifting our demand observation $D_t$ at time $t$ to $D_t - \Theta^{d} p_t$, where $\Theta^d$ is a diagonal matrix consisting of the previously learned diagonal entries of $\Theta$. 
\end{remark}

\subsection{Pricing with Optimism in the Face of Uncertainty}
We design our pricing policies based on the optimism-in-the-face-of-uncertainty (OFU) approach used in the bandit literature \cite{abbasi2011improved}. Specifically, we seek to maintain \textit{confidence sets} $C_t$ that contain the optimal $\Theta^{*}$ with high probability. These sets are updated in an online fashion, using the history $H_t = (D_1, p_1, \ldots, D_{t-1}, p_{t-1})$ of product prices and demand responses such that given any fixed threshold $\delta > 0$, the confidence set $C_t$ contains $\Theta^{*}$ with probability $1 - \delta$. The policy then chooses an optimistic price $p_t$ as follows:
\begin{equation*}
    (p_t, \tilde{\Theta}_t) = \text{arg}\max_{\Theta \in C_t, p \in \mathcal{P}} \{r_{\Theta}(p)\}.
\end{equation*}
In what follows, we focus primarily on approaches for constructing the confidence sets $C_t$ under the different sparsity frameworks described above. 

\subsection{PAC-Bayesian Estimator}
We derive our estimator $\hat{\Theta}_t$ by applying a PAC-Bayesian approach with sparsity priors \cite{alquier2011pac, alquier2015bayesian}. At each iteration, we sample $\hat{\Theta}_t$ over a prior, which ``encourages'' $\hat{\Theta}_t$ to exhibit a desired sparsity constraint. We first define the least-squares empirical risk as
\begin{equation}
\label{eq: ERM}
    r(\Theta) =  \frac{1}{t}\sum_{s = 1}^{t} \|D_s - \Theta p_s\|^2.
\end{equation}
(In this subsection, we occasionally suppress the dependence on $t$ for notational simplicity.)
We construct our sparsity prior in two-stages.
\begin{enumerate}[leftmargin = *]
    \item \textit{Sparsity}: Our sparsity prior has the form
    \begin{equation}
    \label{Sparsity Prior}
        \mu(\Theta) = \sum_{J \in \mathcal{J}} \pi_J \mu_J(\Theta),
    \end{equation}
    where $\mathcal{J}$ is some (possibly uncountable) family of spaces with specific sparsity constraints, $\nu_J(\Theta)$ is a measure over $J$ (often uniform), and $\pi_J$ are the mixing probabilities. In Section \ref{Technical Contributions}, we discuss various parameterizations of sparsity under our different frameworks---the members of the family $\mathcal{J}$ are parameterized accordingly. Then, our mixing probabilities $\pi_J$ are designed to skew towards low values of the sparsity parameter. 
    \item \textit{Exponential Weights}: Let $Z(\cdot)$ be a scaling operator defined on the $\text{supp}(\mu)$. Then, we define $\rho_{\lambda}$ to be the measure:
    \begin{equation}
    \label{Exponential Weights}
        \frac{d\rho^{\lambda}_t(\Theta)}{d\mu(\Theta)} = \frac{e^{-\lambda r(Z(\Theta))}}{\int_{\Theta} e^{-\lambda r(Z(\Theta))}\mu(\Theta)d\Theta},
    \end{equation}
    where $\lambda > 0$ is a parameter. The exponential density of $\rho_{\lambda}$ encourages the prior to minimize the empirical risk in \eqref{eq: ERM}.
    Observe that $\rho_{\lambda}$ implicitly depends on $t$ through the empirical risk $r(\Theta)$ and $\lambda$, which can vary over time. We compute our estimator $\hat{\Theta}$ as the expected value of $Z(\Theta)$ with respect to $\rho_{\lambda}$; i.e.,
\begin{equation}
\label{Exponential Weights}
    \hat{\Theta} = \int Z(\Theta) d\rho^{\lambda}_t(\Theta).
\end{equation}
\end{enumerate}

\begin{remark}[Notation]
In the following, we will denote the true price-sensitivity matrix as $\Theta^{*}$ to emphasize that it is the parameter to be learned. For a matrix $A$, we denote by $||A||$ its operator norm and $||A||_{\text{Fro}}$ its Frobenius norm. For any two Banach spaces $\mathcal{B}_1$ and $\mathcal{B}_2$, $||\cdot||_{\theta, p}$ denotes the norm in the interpolation space $[\mathcal{B}_1, \mathcal{B}_2]_{\theta, p}$ (when the ambient spaces are understood from context). Moreover, $||\cdot||_{\mathcal{H}^{\beta}}$ denotes the norm on $||\cdot||_{\mathcal{H}^{\beta}_{\mathcal{K}}}$ when the kernel $\mathcal{K}$ is understood, and by $||\cdot||_{\mathcal{K}}$ the norm in $\mathcal{H}_{\mathcal{K}}$ (with reproducing kernel $\mathcal{K}$). We use $\mathcal{D}_{\text{KL}}(\cdot | \cdot)$ to denote the KL-divergence, and $\mathcal{M}(\mu)$ to denote the set of measures absolutely continuous to $\mu$. Finally, we say $f(x) = \mathcal{O}(g(x))$ if $f(x) \leq Cg(x)$ for sufficiently large $x$ and $f(x) \asymp g(x)$ if $\lim_{x \to \infty} \frac{f(x)}{g(x)}  =  1$. 
\end{remark}
\section{Analysis} \label{Technical Contributions}
We first discuss the conversion of confidence sets $C_t$ to regret bounds via the OFU policy. Suppose that we are given an online algorithm to construct $C_t$. Based on this, the following result demonstrates how the radius of the confidence sets translates into an upper bound on the pseudo-regret after $T$ iterations.
\begin{lemma}
\label{Confidence Set to Regret}
For $t = 1,2,\ldots,$ let $V_t = \sum_{s=1}^t p_s^{} p_s^\intercal$, and define:
\begin{equation*}
    C_t = \{\Theta: {\sf tr}((\Theta - \hat{\Theta}_t)^\intercal V_t (\Theta - \hat{\Theta}_t)) < \beta_t^2(\epsilon), ||\Theta|| \leq K\}.
\end{equation*}
Suppose that $\Theta^{*} \in C_t$ for $t = 1,2,\ldots,$ with probability at least $1 - \epsilon$, and $||\Theta^{*}||, ||\hat{\Theta}|| \leq K$ Then, we have with probability at least $1 - \epsilon$ that
\begin{equation}
\label{C-R Formula}
    \Delta_T \leq L^2\sqrt{8N\log \left(1 + \frac{2TL^2}{N}\right)}\sqrt{\sum_{t = 1}^T \beta_t^2(\epsilon) + 2K^2TN}.
\end{equation}
\end{lemma}
The proof of Lemma \ref{Confidence Set to Regret} is in the supplementary materials. The bound in \eqref{C-R Formula} illustrates that our best-case regret is $\mathcal{O}(\sqrt{NT})$. Hence, in order to preserve the asymptotically optimal $\sqrt{T}$ rate, we need to ensure that $\beta_t^2(\epsilon)$ is at worst $\mathcal{O}(T)$. In Sections \ref{L_0 Sparsity} and \ref{Spectral Sparsity}, we outline the measures $\{\mu_J\}_{J \in \mathcal{J}}$ used to achieve this aim in different sparsity frameworks. Moreover, the bound $K$ is known by the seller and may depend on $N$ --- in the remainder of the paper, we discuss suitable growth rates of $K$ in our different sparsity contexts. We first present a general result that provides a variational characterization of the radius of the confidence sets $C_t$:
\begin{lemma}
Suppose $||\Theta||_{2} \leq K$ and $Z(\Theta) = \frac{K\Theta}{\|\Theta\|}$. Define:
\begin{equation*}
    R(\Theta) = \frac{1}{t}\sum_{s = 1}^t ||(\Theta - \Theta^{*})p_s||^2.
\end{equation*}
Then, letting $\lambda = \frac{t}{2\mathcal{C}_1}$ in \eqref{Exponential Weights}, we have:
\label{Variational Ellipsoid}
\begin{equation}
\label{Confidence Set Formula}
    \frac{1}{t}\sum_{s=1}^t \|(\hat{\Theta}_t - \Theta^{*}) p_s\|^2 \leq \inf_{\rho \in \mathcal{M}(\mu)} 3\int R(Z(\bar{\Theta})) d\rho(\bar{\Theta}) + \frac{8\mathcal{C}_1}{t}\Big(\mathcal{D}_{\text{KL}}(\rho||\mu) + \log \frac{2}{\epsilon} \Big)
\end{equation}
for $t = 1, 2, \ldots$ with probability $1 - \epsilon$.
\end{lemma}
The absolute constant $\mathcal{C}_1$ appearing in \eqref{Confidence Set Formula} is defined as:
\begin{equation*}
    \mathcal{C}_1 = \max\left\{(Q + KL)KL,\, \sigma^2 + K^2\right\}.
\end{equation*}
which will be used repeatedly throughout the sequel. We observe that \eqref{Confidence Set Formula} precisely characterizes our target confidence set, as $\sum_{s=1}^t \|(\hat{\Theta}_t - \Theta^{*})p_s\|^2 = {\sf tr}((\hat{\Theta}_t - \Theta)^\intercal V_t(\hat{\Theta}_t - \Theta))$ with $V_t$ being defined as in Lemma \ref{Confidence Set to Regret}. Thus, the right-hand side of \eqref{Confidence Set Formula} gives us the desired radius $\beta_t^2(\delta)$ of $C_t$ to plug into Lemma \ref{Confidence Set to Regret}. In the remainder of the paper we customize and synthesize the above two results over a variety of problem settings. 
\subsection{Analysis of $L_0$ Sparsity} \label{L_0 Sparsity}
In our first setting, we use the sparsity measure inspired by that presented on p.131 in \cite{alquier2011pac}. That is, for each $J \subset [N] \times [N]$, we define the truncated set $J_K$ of matrices $\Theta$ supported on $J$ as
\begin{equation*}
    J_{K} = \{\Theta \in \mathbb{R}^{N \times N}: \|\Theta\|^2_{\text{Fro}} \leq K^2 N + 1,\, \Theta_{ij} \neq 0 \text{ if and only if } (i, j) \in J\}
\end{equation*}
and our mixing probabilities as 
\begin{equation}
\label{Full-Sparse Mixing}
\pi_{J} = \frac{\alpha^{|J|}}{\sum_{i = 0}^{N^2} \alpha^i} {N^2 \choose |J|}^{-1} 
\end{equation}
for some $\alpha \in (0,1)$. Hence, we let our family of support spaces be $\mathcal{J} = \{J_K: J \subset [N] \times [N]\}$, choose $\mu_{J_K}$ to be the uniform spherical measure over $J_K$ i.e up to isomorphism:
\begin{equation*}
    \mu_{J_K}(\cdot) = \frac{\mathbbm{1}_{\{\|\Theta\|_{\text{Fro}} \leq K^2 N + 1\}}m_{J}(\cdot)}{m_{J}(\sqrt{K^2 N +1}\mathbb{B}_{J})},
\end{equation*}
where $m_{J}(\cdot)$ is the Lebesgue measure over $\mathbb{R}^{|J|}$ and $\mathbb{B}_{J}$ is the unit Euclidean ball in $\mathbb{R}^{|J|}$. Finally, we set $Z(\Theta) = \frac{K\Theta}{||\Theta||_2}$, and then define the sparsity prior $\mu(\Theta)$ using \eqref{Sparsity Prior}. Then applying this prior with Lemma \ref{Confidence Set Formula} to our online setting, we obtain the following result.
\begin{lemma}
\label{Adapted Case}
Suppose that $\|\Theta^{*}\| \leq K$, and let $J^{*} = \{(i, j) \in [N] \times [N]: \Theta^{*}_{ij} \neq 0\}$ denote the support of $\Theta^{*}$. Then, setting $\lambda_t = t/(2\mathcal{C}_1)$ in \eqref{Sparsity Prior}, we have
\begin{align}
\label{eq: Full-Sparse Confidence}
   \sum_{s=1}^t & \|(\hat{\Theta}_t - \Theta^{*}) p_s\|^2 \notag \\ &\leq \frac{3L^2}{t} + 8\mathcal{C}_1\Big(|J^{*}|\log{\Big(\frac{eN^2 t\sqrt{K^2 N + 1}}{|J^{*}|}\Big)} - \log{\alpha(1 - \alpha)}  + \log \frac{2}{\epsilon} \Big)
\end{align}
for $t = 1,2,\ldots,$ with probability at least $1 - \epsilon$.
\end{lemma}

Equipped with the full description of the confidence set, we can now apply Lemma \ref{Confidence Set to Regret} and the estimate $\sum_{t = 1}^T \log t \leq \int_{1}^{T + 1} \log t dt$ to obtain our regret bound. We omit the full proof the regret bounds in Theorem \ref{Full-Sparse Regret} and the sequel as these follow directly from the confidence radius in \eqref{eq: Full-Sparse Confidence}, Lemma \ref{Confidence Set to Regret} and the straightforward estimation of the sum $\sum_{t = 1}^T \beta_t^2(\epsilon)$ in \eqref{C-R Formula} by the corresponding integral as described previously:
\begin{theorem}
\label{Full-Sparse Regret}
Suppose that the assumptions of Lemma \ref{Adapted Case} hold and $K = \mathcal{O}(\log^{a} N)$ for $a > 1$. Then, we have
\begin{small}
\begin{align*}
    & \Delta_T \\
    & \leq  L^2 \sqrt{3L^2 \log T + 8\mathcal{C}_1 T\left[|J^{*}|\log{\Big(\frac{eN^2\sqrt{K^2 N + 1}}{|J^{*}|}\Big)} - \log{\alpha(1 - \alpha)} + \log \frac{2}{\epsilon}\right] + 8\mathcal{C}_1 |J^{*}| (T + 1)\log (T + 1) + 2TN\log^{2a}(N)} \\ 
    & \hspace{20pt} \cdot \sqrt{8N\log \left(1 + \frac{2TL^2}{N}\right)}
\end{align*}
\end{small}
for $T = 1,2,\ldots,$ with probability at least $1 - \epsilon$.
\end{theorem}

\par Our choice to set $K = \mathcal{O}(\log^{a} N)$ above is motivated by the theory of sparse random graphs. Indeed, suppose our product network is a realization of an Erdos-Renyi graph with edge probabilities $\frac{\log N}{N}$ and uniformly bounded edge weights, i.e. the probability of any two products being connected decreases with the number of products, while the total number of connections grows logarithmically. Then, Theorem 1.1 in \cite{krivelevich2003largest} gives us that $||\Theta^{*}|| = \mathcal{O}(\log^2 N)$ when $\Theta^{*}$ is interpreted as the adjacency matrix of the product network. 

\par Our regret bound in Theorem \ref{Full-Sparse Regret} is $\mathcal{O}(\sqrt{|J^{*}|NT})$ up to logarithmic terms. In the worst case, i.e. the when $\Theta^{*}$ is full and $|J^{*}| = N^2$,  we obtain a rate that is  $\mathcal{O}(\sqrt{N^{3}T})$. While this rate is asympotically optimal in $T$,  the linear dependence on $|J^{*}|$ is not ideal, as all nonzero price-sensitivities may not be equally influential. In the remainder of this paper, we will examine how the scalability of our algorithm in $N$ can be improved under more specified scenarios, where the assumption of spectral sparsity is replaced by more \textit{a priori} knowledge of the structure of the product network.

\subsection{Analysis of Off-Diagonal Sparsity} \label{Specified Sparsity}
\par In our first such specified setting, we suppose the weights on the diagonal of $\Theta$ dominate those on the off-diagonal by an order of magnitude. As alluded to earlier, this indicates that a product's price influences its own demand far more than the demand of any other product. Specifically, we assume that the off-diagonal entries of $\Theta$ are $\mathcal{O}(1/N)$, while the diagonal entries of $\Theta$ are $\mathcal{O}(1)$ as $N$ grows large; i.e., as the number of products increases, individual cross-price sensitivities are diminished while own-price sensitivities are preserved. 

\par Note that in this setting, we do not require the true price-sensitivity matrix $\Theta^{*}$ to have zero entries. Indeed, we will demonstrate that our policy produces $\mathcal{O}(N\sqrt{T})$ regret even when $\Theta^{*}$ is full. Moreover, in this setting our assumption $||\Theta^{*}|| \leq K$ is virtually equivalent to simply bounding the diagonal entries of $\Theta^{*}$. Indeed, since the off-diagonal entries of $\Theta^{*}$ are $\mathcal{O}\Big(\frac{1}{N}\Big)$, by applying the Gershgorin circle theorem to $\Theta^{*}(\Theta^{*})^T$ (which exhibits the same property), $||\Theta^{*}|| \leq \max_{i \in [N]} \Theta^{*}_{ii} + \mathcal{O}(1)$. Meanwhile the diagonal entries of $\Theta^{*}$ are clearly bounded by $||\Theta^{*}||$. 

\paragraph{} In this new setting, our previous prior would no longer be ideal because, although it encourages sparsity, it does not differentiate between the diagonal and off-diagonal entries of $\Theta$. To address these localizations, we simply modify the measures $\mu_{J}$ and mixing probabilities $\pi_{J}$, and leave the remaining structure of the prior (i.e., the family of spaces $\mathcal{J} = \{J: J \subset [N] \times [N]\}$) unchanged. 
\begin{align}
    \Theta_{ii} & \sim |N(0, 1)| \hspace{2mm} \forall (i, i) \in J, \label{eq: DiagSparse} \\
    \Theta_{ij} & \sim \sqrt{\Gamma(N^{-1}, 1)} \hspace{2mm} \forall (i, j) \in J \text{ with } i \neq j. \label{eq: OffDiagSparse}
\end{align}
where $\Gamma(k, \theta)$ denotes a gamma distribution with shape $k$ and scale $\theta$ (note we allow $\Theta_{ij}$ to take both positive and negative values with equal probability). Note that $|N(0, 1)| \sim \sqrt{\Gamma(0.5, 2)}$ by definition, so the diagonal entries in \eqref{eq: DiagSparse} are indeed related to the off-diagonal ones in \eqref{eq: OffDiagSparse} only differing in their scale and shape (and possibly their sign). We then set
\begin{equation}
\label{Specified Sparsity Mixing}
    \pi_{J} = \frac{|J|^{-2}}{\sum_{i = 0}^{N^2} i^{-2}} {N^2 \choose |J|}^{-1}.
\end{equation}
\paragraph{} Like the mixing probabilities in \eqref{Full-Sparse Mixing}, those in \eqref{Specified Sparsity Mixing} vary inversely with the size of the support set $J$. However, the variation is less drastic in \eqref{Specified Sparsity Mixing}, with the probabilities decaying only polynomially in the $|J|$ rather than exponentially --- this difference will be crucial for the scalability of our regret in Theorem \ref{Orders-Regret}. Finally, we set our scaling $Z = \frac{K\Theta}{||\Theta||}$, and as before combine these components to construct our sparsity prior $\mu(\cdot)$ according to \eqref{Sparsity Prior} and \eqref{Exponential Weights}. We now present the confidence set for this setting:
\begin{lemma}
\label{Orders Ellipsoid}
Suppose that $\Theta^{*}_{ij} \leq \frac{C}{N}$ for some $C > 0$ and all $i,j$ satisfying $i \neq j$, in addition to the assumptions of Lemma \ref{Adapted Case}. Let $\theta = \min{\{|\Theta^{*}_{ij}|, \Theta^{*}_{ij} \neq 0\}} $. Then, using the prior defined in \eqref{eq: DiagSparse}, \eqref{eq: OffDiagSparse}, and \eqref{Specified Sparsity Mixing}, and setting $\lambda = t/(2\mathcal{C}_1)$ in \eqref{Exponential Weights}, we have for $N > \frac{3C}{\sqrt{2}}$,
\begin{equation}
\label{Orders Intermediate}
    \sum_{s=1}^t \|(\hat{\Theta}_t - \Theta^{*})p_s\|^2 \leq \frac{3L^2\theta^2}{t} + 8\mathcal{C}_1\left[K^2N + (2N -1)\log \theta^{-1}t + 4 \log N + \log{\frac{\pi^2}{6}} + \log \left(\frac{2}{\epsilon}\right)\right]
\end{equation}
for $t=1,2,\ldots,$ with probability at least $1 - \epsilon$.
\end{lemma}
Substituting the radius in \eqref{Orders Intermediate} into the regret bound from Lemma \ref{Confidence Set to Regret}, we obtain:
\begin{theorem}
\label{Orders-Regret}
Suppose that the assumptions of Lemma \ref{Orders Ellipsoid} hold. Then, we have
\begin{align*}
    \Delta_T & \leq L^2\sqrt{3L^2\theta^2 \log T + 8\mathcal{C}_1T\left[K^2N + CN \log \theta^{-1} T + \log \left(\frac{2}{\epsilon}\right)\right] + 2K^2TN} \\
    &\hspace{20pt} \cdot \sqrt{8N\log \left(1 + \frac{2TL^2}{N}\right)}
\end{align*}
for $T=1,2,\ldots,$ with probability at least $1 - \epsilon$ (where $C$ is an absolute constant)
\end{theorem}
\paragraph{} In Theorem \ref{Orders-Regret}, we observe that we preserve the $\mathcal{O}(\sqrt{T})$ growth of regret in $T$ from Theorem \ref{Full-Sparse Regret}, yet the growth of regret is now linear in $N$, the number of products, as opposed to $|J^{*}|$, the number of connections in the graph. This can be attributed to the fact that in the setting of Theorem \ref{Orders-Regret}, we maintain that while cross-product price effects may exist, they are not as influential as the relationship between a product's own price and its demand. The $\mathcal{O}(N)$ growth of regret is optimal in this setting, because this is the best achievable growth rate of regret even in the absence of cross-product price effects. To see this, consider $N$ single-product pricing problems, invoke the lower bound on regret in Theorem 1 of \cite{keskin2014dynamic} in each problem, and add up the $N$ lower bounds.

\subsection{Analysis of Spectral Sparsity} \label{Spectral Sparsity}

\par In this section, we demonstrate that we may achieve a $\mathcal{O}(N\sqrt{T})$ regret rate even when $\Theta^{*}$ is dense, under certain smoothness assumptions on the price-sensitivity function $\kappa^{*}(x_i, x_j) = \Theta^{*}_{ij}$. We study two PAC-Bayesian priors, which simulate the generating function $\kappa$ using a family of Gaussian processes. In our first prior, this family consists of different scalings of a single Gaussian process $\text{GP}(0, \mathcal{K})$ with covariance kernel $\mathcal{K}$ and corresponding RKHS $\mathcal{H}_{\mathcal{K}}$. While this prior requires minimal structural information on the process  $\text{GP}(0, \mathcal{K})$, it only produces the optimal $\mathcal{O}(N\sqrt{T})$ regret rate in the correctly specified case (when $\kappa^{*} \in \mathcal{H}$) and is not robust to misspecification. Our second prior simulates $\kappa$ over the different \textit{powers} $\{\text{GP}(0, \mathcal{K}^{\gamma})\}_{\gamma \in (0, 1)}$ of the process  $\text{GP}(0, \mathcal{K})$. Implementing this prior requires spectral information on the process $\text{GP}(0, \mathcal{K})$ making its application more limited than the scaling prior --- however, we will see that this prior produces the optimal $\mathcal{O}(N\sqrt{T})$ regret rate in both the correctly specified and misspecified settings (when $\kappa^{*} \not \in \mathcal{H}$). 

\par We note that in this setting, our assumption that $||\Theta^{*}|| \leq K$ stems naturally from the smoothness of the generating function $\kappa^{*}$. Indeed, for a simple example, suppose $\Omega$ is isomorphic to $[-\pi, \pi]^d$ ($d \in \mathbb{N}$) and $\kappa^{*}$ is an isotropic $C^{\infty}$ positive-definite function. Then, if the products are distributed uniformly over $\Omega$, the eigenvalues of $\Theta^{*}$ approximate those of the integral operator $T_{\mathcal{L}}$ induced by the kernel $\mathcal{L}(x, y) = \kappa^{*}(x - y)$, for sufficiently large $N$. Since the Fourier coefficients of a $C^{\infty}$ function decay faster than any polynomial, by Theorem  1 of \cite{reade1992eigenvalues} the same follows for $T_{\mathcal{L}}$ and hence $\Theta^{*}$, implying that the eigenvalues are clearly bounded.  

\subsubsection{Scaling Prior} 
\par We recall that our spectral sparsity assumption concerns the smoothness of our target function $\kappa$. First, we study a scaling prior, inspired by \cite{suzuki2012pac}, in which our parameter of interest $\gamma \in (0, \infty)$ scales the Gaussian process with RKHS $\mathcal{H}$. From the Feldman-Hajek theorem, different scalings of a given Gaussian process induce mutually singular measures --- we mix these measures using the mixing probabilities $\pi(\gamma)$:
\begin{align}
    \mu_{\gamma}: \kappa(\cdot) & \sim \text{GP}\left(0, \gamma K\right), \label{eq: GP Scaling} \\
    \mu & = \int_{1}^{\infty} \pi(\gamma)\mu_{\gamma}d\gamma, \nonumber 
\end{align}
where $\pi$ is some probability density on $[1, \infty)$. We again set our scaling operator to $Z(\Theta) = \frac{K\Theta}{||\Theta||}$. Note in both the proofs and discussion for this section, we often abuse notation and interchangeably refer to a random matrix or function with distribution $\mu_{\gamma}$ --- it should be understood that in these cases the matrix $\Theta$ and function $\kappa$ are related by $\Theta_{ij} = \kappa(g_i - g_j)$. We first present the following characterization of our confidence sets:

\begin{lemma}
\label{Spectral Scaling Confidence, Exponential}
Suppose that $\|\Theta^{*}\| \leq K = \mathcal{O}(1)$, $\mu_i \asymp e^{-iq}$ for $q > 0$, and $\mathcal{H}^{\beta} \hookrightarrow L^{\infty}$ continuously for some $0 < \beta < \alpha$. Then, setting $\lambda_t = t/(2\mathcal{C}_1)$, we have:
\begin{equation}
\label{Spectral Scaling Intermediate, Exponential}
    \sum_{s=1}^t \|(\hat{\Theta}_t - \Theta^{*})p_s\|^2 \leq 3L^2 + 8\mathcal{C}_1\Big(\mathcal{C}_{\alpha, \beta}||\kappa^{*}||_{\frac{\alpha - \beta}{1 - \beta}, \infty}^{\frac{2(1 - \beta)}{\alpha - \beta}}(4N^2t)^{\frac{1 - \alpha}{1 - \beta}} + \mathcal{C}_{\beta, q}\log^2  \Big(2N\sqrt{t}\Big) + \log \frac{2}{\epsilon} \Big)
\end{equation}
for $t=1,2,\ldots,$ with probability at least $1 - \epsilon$, where $\mathcal{C}_{\beta, q}$ depends only on $\beta, q$ (and likewise for $\mathcal{C}_{\alpha, \beta}$).
\end{lemma}

The condition $\mathcal{H}^{\beta} \hookrightarrow L^{\infty}$ continuously for some $0 < \beta < \alpha$ ensures that when $\kappa \sim \mu_{\gamma}$ ($\gamma > \beta$), we can estimate the concentration of $||\kappa - \kappa^{*}||_{\infty}$ via the Hilbert space norm $||\kappa - \kappa^{*}||_{\mathcal{H}^{\beta}}$, which can studied directly via the eigenvalues $\mu_i$ of $T_{\mathcal{K}}$. The proof of Lemma \ref{Spectral Scaling Confidence, Exponential} centers on handling misspecification by approximating $\kappa^{*} \in \mathcal{H}^{\alpha}$ by an element $g \in \mathcal{H}$, contained in the shared Cameron-Martin RKHS $\mathcal{H}$ of the scaled Gaussian measures $\text{GP}\left(0, \gamma \mathcal{K}\right)$, and hence approximating the shifted measures $\text{GP}\left(\kappa^{*}, \gamma \mathcal{K}\right)$ via $\text{GP}\left(g, \gamma \mathcal{K}\right)$, whose behavior is well-studied. Note, however the distribution of the latter shifted measure depends on $||g||_{\mathcal{K}}$, whose magnitude we can only control polynomially in the approximation error $||g - \kappa^{*}||^{-1}_{\mathcal{H}^{\beta}} \asymp \sqrt{t}$ (since we wish to approximate $\kappa^{*}$ more strongly as $t \to \infty$). This by-product of our approximation technique introduces the  term $(4N^2t)^{\frac{1 - \alpha}{\alpha - \beta}}$ in \eqref{Spectral Scaling Intermediate, Exponential} which may be large if $1 - \alpha$ is large and the problem is highly misspecified (note this term vanishes in  the correctly specified situation when $\alpha = 1$). As we observe in Theorem \ref{Spectral Scaling Regret, Exponential}, this may lead to a suboptimal regret rate when we integrate the confidence radius over our horizon $[0, T]$ to obtain our regret bound. Indeed, applying Lemma \ref{Confidence Set to Regret} again to the RHS of \eqref{Spectral Scaling Intermediate, Exponential} with the estimate $\sum_{t = 1}^{T} \log^2(2N\sqrt{t}) \leq \sum_{t = 1}^{T} \log^2(2Nt) \leq  2\sum_{t = 1}^{T} (\log^2(2N) + \log^2 t) \leq 2\log^2(2N)T + 2\int_{1}^{T + 1} \log^2 (t) dt  \leq 2T\log^2 (2N) + 2(T+1)\log^2(T + 1) \leq 2T\log^2 (8NT)$, we obtain the regret bound:

\begin{theorem}
\label{Spectral Scaling Regret, Exponential}
Suppose that the assumptions of Lemma \ref{Spectral Scaling Confidence, Exponential} hold. Then, we have:
\begin{align}
\label{Spectral Scaling Regret Formula, Exponential}
    \Delta_T & \leq L^2\sqrt{3L^2 T + 8\mathcal{C}_1T\left[\mathcal{C}_{\alpha, \beta, \kappa^{*}}(4N^2 T)^{\frac{1 - \alpha}{1 - \beta}} + \mathcal{C}_{\beta, q}\log^2 \Big(8NT\Big) + \log \frac{2}{\epsilon}\right] + 2K^2TN} \\
    & \hspace{20pt} \cdot \sqrt{8N\log \left(1 + \frac{2TL^2}{N}\right)} \nonumber
\end{align}
for $t=1,2,\ldots,$ with probability at least $1 - \epsilon$, where $\mathcal{C}_{\alpha, \beta, \kappa^{*}}$ depends only on $\alpha, \beta$, and $\kappa^{*}$ (and likewise for $\mathcal{C}_{\beta, q}$).
\end{theorem}

Hence, we obtain an adaptive $\mathcal{O}(N^{\frac{1}{2} + \max\{\frac{1}{2}, \frac{1 - \alpha}{1 - \beta}\}}T^{\frac{1}{2} + \frac{1 - \alpha}{1 - \beta}})$ regret rate. Since $\alpha < \beta$ by assumption, we observe that in the worst case this rate becomes $\mathcal{O}(N^{\frac{3}{2}} T)$. While the former rate is clearly not optimal for $\alpha \neq 1$ (i.e. under misspecification), we emphasize again, that in the correctly specified setting (where $\alpha = 1$) the approximation term $(4N^2 T)^{\frac{1 - \alpha}{1 - \beta}}$ vanishes and we obtain the optimal rate of $\mathcal{O}(N\sqrt{T})$. In the next section, we observe how we can eliminate this polynomial term in the misspecified setting as well, when we have closed form characterizations of the interpolation spaces. 

\subsubsection{Powers Prior}
\par In several settings, including the popular square-exponential (Gaussian) kernels, closed form expressions are available for the intermediate spaces $\mathcal{H}^{\gamma}$ ($\gamma \in (0, 1)$) and/or their kernels $K^{\gamma}$ (see e.g section 5 in \cite{karvonen2021small} for further examples). In these scenarios, we can simulate our Gaussian measures directly over these interpolation spaces, and thereby approximate $\kappa^{*}$ directly in its native space $\mathcal{H}^{\alpha}$. Here our family of supports is $\mathcal{J} = \{\text{supp}(\mu_{\gamma}): \gamma \in (0, 1] \}$, where $\mu_{\gamma}$ consists of the centered Gaussian measure with covariance $K^{\gamma}$ whose support $\text{supp}(\mu_{\gamma})$ is itself contained in an RKHS power space under suitable assumptions (see \cite{steinwart2019convergence} for a comprehensive discussion). Recall $K^{\gamma}$ is the kernel of the interpolation spaces $\mathcal{H}^{\gamma}$, which is reproducing iff it is almost surely finite. We define our sparsity prior:
\begin{align}
    \mu_{\gamma}: \kappa(\cdot) & \sim \text{GP}\left(0, K^{\gamma}\right), \label{eq: GP Powers} \\
    \mu & = \int_{0}^1 \pi(\gamma)\mu_{\gamma}d\gamma. \nonumber 
\end{align}
Here, $\pi(\gamma)$ is an arbitrary mixing density on $(0, 1]$ and again $Z(\Theta) = \frac{K\Theta}{||\Theta||}$. We first present the confidence sets induced by this prior:

\begin{lemma}
\label{Spectral Confidence, Exponential}
Suppose that $\|\Theta^{*}\| \leq K = \mathcal{O}(1)$ and $\mu_i \asymp e^{-iq}$, and $\mathcal{H}^{\beta} \hookrightarrow L^{\infty}$ continuously for some $0 < \beta < \alpha$. Then, setting $\lambda_t = t/(2\mathcal{C}_1)$ we have:
\begin{equation}
\label{Spectral Intermediate Exponential}
    \sum_{s=1}^t \|(\hat{\Theta}_t - \Theta^{*})p_s\|^2 \leq \frac{3L^2}{t} + 8\mathcal{C}_1\Big(\mathcal{C}_{\alpha, \beta}||\kappa^{*}||^2_{\mathcal{H}^{\alpha}} + \mathcal{C}_{\beta, q, \alpha}\log^2 (tN^2) + \log \frac{2}{\epsilon} \Big)
\end{equation}
for $t=1,2,\ldots,$ with probability at least $1 - \epsilon$, where $\mathcal{C}_{\alpha, \beta, q}$ depends only on $\alpha, \beta$ and $q$ (and likewise for $\mathcal{C}_{\alpha, \beta}$).
\end{lemma}

Observe that in \eqref{Spectral Intermediate Exponential} we have eliminated the approximation error producing the polynomial term in \eqref{Spectral Scaling Intermediate, Exponential} as our powers prior in \eqref{eq: GP Powers} allows us to approximate $\kappa^{*}$ directly in $\mathcal{H}^{\alpha}$. Thus, our confidence radius in \eqref{Spectral Intermediate Exponential} grows only polylogarithmically in $N$ and $T$. It is important to note that with the powers prior in \eqref{eq: GP Powers}, the smoothness of the sample paths of $\text{GP}(0, K^{\gamma})$ varies with $\gamma$, while in the scaling prior of \eqref{eq: GP Scaling}, the Cameron-Martin space of $\mu_{\gamma}$ is fixed (up to norm equivalence). Now again applying Lemma \ref{Confidence Set to Regret} and summing over the confidence set radii in \eqref{Spectral Intermediate Exponential}, we obtain the following regret bound:

\begin{theorem}
\label{Spectral Regret, Exponential}
Suppose that the assumptions of Lemma \ref{Spectral Confidence, Exponential} hold. Then, we have
\begin{small}
\begin{align}
\label{Spectral Regret Formula, Exponential}
    \Delta_T & \leq L^2\sqrt{3L^2 \log T + 8\mathcal{C}_1T\left[\mathcal{C}_{\alpha, \beta}||\kappa^{*}||^2_{\mathcal{H}^{\alpha}} + \mathcal{C}_{\beta, q, \alpha}(\log^2 (N^2) + \log^2 T) + \log \frac{2}{\epsilon}\right] + 2K^2TN} \\
    & \hspace{20pt} \cdot \sqrt{8N\log \left(1 + \frac{2TL^2}{N}\right)} 
\end{align}
\end{small}
for $T=1,2,\ldots,$ with probability at least $1 - \epsilon$. 
\end{theorem}

Observe that in Theorem \ref{Spectral Regret, Exponential}, the degree of misspecfication (i.e. the value of $\alpha$) only manifests in the scalars $||\kappa^{*}||^2_{\mathcal{H}^{\alpha}}$ and $\mathcal{C}_{\beta, p, \alpha}$ and hence does not affect the asymptotic behavior of the bound in either $N$ or $T$. Hence, even in the misspecified case, using the powers prior of \eqref{eq: GP Powers} produces a regret bound that is worst case $\mathcal{O}(N\sqrt{T})$ (up to polylogarithmic terms).

\section{Discussion} \label{Discussion}
In this paper, we explore a novel framework for analyzing dynamic pricing with demand learning on large product networks. By combining online bandit algorithms with PAC-Bayesian and Gaussian process methods, we develop a new approach for treating sparsity and achieving optimal regret rates. There are several remaining questions that we hope to explore in future analyses. Namely, in this setting, we considered the problem of learning the coefficients matrix with a known intercept term in the linear demand model. In future work, we hope to extend our framework to learning both the intercept $d_0$ and parameter $\Theta$ simultaneously. We anticipate that this adjustment will not affect the regret rates, however this requires a further analysis. Finally, in subsequent work, we also hope to consider nonlinear demand models and nonparametric noise settings.

%
%

\section{Proofs}
\begin{proof}{Proof of Lemma \ref{Confidence Set to Regret}.}
Let $\lambda > 0$, $\bar{V}_t = \lambda I + \sum_{i = 1}^t p_i p_i^T$, and denote $r_t = r_{\Theta^{*}}(p^{*}) - r_{\Theta^{*}}(p_t)$:
\begin{align*}
    r_{\Theta^{*}} &= \phi(\Theta^{*}, p^{*}) - \phi(\Theta^{*}, p_t) \\
    & \leq \phi(\hat{\Theta_t}, p_t) - \phi(\Theta^{*}, p_t)  \\
    & = p^T_t  (\hat{\Theta_t} p_t + d_0)  - p_t^T(\Theta^{*} p_t + d_0)\\
    & =  p^T_t (\hat{\Theta_t} - \tilde{\Theta}_{t-1}) p_t + p_t^T (\tilde{\Theta}_{t-1} - \Theta^{*})p_t.
\end{align*}
Then, applying Cauchy-Schwarz with respect to the norm $||\cdot||_{\bar{V}_{t-1}}$, we have, with probability $1 - \epsilon$ for all $t \geq 1$:
\begin{align*}
    r_t & \leq L||\tilde{\Theta_t} - \Theta_{t-1}||_{\bar{V}_{t-1}} ||p_t||_{\bar{V}^{-1}_{t-1}} + L||\Theta_{t-1} - \Theta^{*}||_{\bar{V}_{t-1}} ||p_t||_{\bar{V}^{-1}_{t-1}}
    & \leq 2L||p_t||_{\bar{V}^{-1}_{t-1}}\sqrt{\beta^2(\delta, t) + \lambda ||\tilde{\Theta}_{t-1} - \Theta^{*}||^2_{\text{Fro}}},
\end{align*}
since $\beta^2(\epsilon, t)$ is the radius of the confidence set. Moreover, modifying the proof of Lemma 11 in \cite{abbasi2012online}, we can show inductively that:
\begin{equation*}
    \text{det}(\bar{V}_t) = (\lambda + ||p_1||^2) \prod_{i = 2}^{t} (1 + ||p_i||_{\bar{V}_{i -1}}^2),
\end{equation*}
from which we obtain (after applying the inequality $\min\{1, x\} \leq 2\log (1 + x)$ as in Lemma 11 from \cite{abbasi2011improved}):
\begin{align}
    \sum_{t = 1}^{T} \min \{1, ||p_t||_{\bar{V}_{t -1}}^2\} & \leq 2 \sum_{t = 1}^{T} \log  (1 + ||p_t||_{\bar{V}_{t -1}}^2) \nonumber \\
    & = 2(\log  (1 + ||p_1||_2^2) - \log  (\lambda + ||p_1||_2^2) + \log \text{det}(\bar{V}_T)) \nonumber \\
    & \leq 2 \log \frac{1 + ||p_1||_2^2}{\lambda + ||p_1||_2^2} + 2N\log \Big(\lambda + \frac{TL^2}{N}\Big), \label{eq: Abbasi Lemma 11}
\end{align}
where we have applied Lemma 10 of \cite{abbasi2012online} in the last step (determinant-trace inequality). Thus, we have that, with probability $1 - \epsilon$:
\begin{align*}
    \sum_{t = 1}^T r_t & \leq 2L \sum_{t = 1}^T ||p_t||_{\bar{V}^{-1}_{t-1}}\sqrt{\beta^2(\epsilon, t) + \lambda ||\tilde{\Theta}_{t-1} - \Theta^{*}||^2_{\text{Fro}}} \\
    & \leq 2L \sqrt{\Big(\sum_{t = 1}^T \beta^2(\epsilon, t) + \lambda ||\tilde{\Theta}_{t-1} - \Theta^{*}||^2_{\text{Fro}}\Big)\Big(\sum_{t = 1}^T ||p_t||^2_{\bar{V}^{-1}_{t-1}}\Big)} \\
    & \leq L^2\sqrt{8\Big(\sum_{t = 1}^T \beta^2(\epsilon, t) + 2K^2 N\lambda \Big)\Big(\sum_{t = 1}^T \log \frac{1 + ||p_1||_2^2}{\lambda + ||p_1||_2^2} + N\log \Big(1 + \frac{TL^2}{N}\Big)\Big)},
\end{align*}
where in the last step, we have noted that $||p_t||^2_{\bar{V}_{t-1}} \leq \frac{L^2}{\lambda}\min\{1, ||p_t||^2_{\bar{V}_{t - 1}}\}$ (since $||p_t||^2_{\bar{V}_{t-1}} \leq \frac{L^2}{\lambda}$ as $||\bar{V}^{-1}_{t-1}||\leq \frac{1}{\lambda}$), applied \eqref{eq: Abbasi Lemma 11} and the fact that $||\tilde{\Theta}_{t-1} - \Theta^{*}||^2_{\text{Fro}} \leq 2\max\{\|\tilde{\Theta}_{t-1}\|^2_{\text{Fro}}, \|\Theta^{*}\|^2_{\text{Fro}}\} \leq 2N\max\{\|\tilde{\Theta}_{t-1}\|^2, \|\Theta^{*}\|^2\} \leq 2K^2N$ by assumption. The result follows from setting $\lambda = 1$.
\end{proof}
\begin{proof}{Proof of Lemma \ref{Variational Ellipsoid}.}
Let $\Theta \in \text{ran}(Z)$ and set $T_i = ||Y_i - \Theta^{*}X_i||^2 - ||Y_i - \Theta X_i||^2$. Then, we have:
\begin{align*}
    \mathbb{E}[||Y_i - \Theta^{*}X_i||^2 - ||Y_i - \Theta X_i||^2 | \mathscr{F}_{i - 1}] & = \mathbb{E}[||W_i||^2 - ||W_i + (\Theta^{*} - \Theta)X_i||^2 | \mathscr{F}_{i - 1}] \\
    & = 2\mathbb{E}[\langle W_i,  (\Theta^{*} - \Theta)X_i \rangle | \mathscr{F}_{i - 1}] - \mathbb{E}[(\Theta^{*} - \Theta)X_i||^2 | \mathscr{F}_{i - 1}] \\
    & = - \mathbb{E}[||(\Theta^{*} - \Theta)X_i||^2 | \mathscr{F}_{i - 1}],
\end{align*}
\begin{align*}
    \mathbb{E}\Big[T^2_i | \mathscr{F}_{i - 1} \Big] & \leq  \mathbb{E}[||2W_i + 2\Theta^{*}X_i - (\Theta + \Theta^{*})X_i||^2 ||(\Theta - \Theta^{*})X_i||^2 | \mathscr{F}_{i - 1} ] \\
    & (8\mathbb{E}[||W_i||^2 | \mathscr{F}_{i - 1}] + 8K^2L^2) \mathbb{E}[||(\Theta - \Theta^{*})X_i||^2 | \mathscr{F}_{i - 1}] \\
    & = (8\sigma^2 + 8K^2L^2) ||(\Theta - \Theta^{*})X_i||^2 \equiv v_i,
\end{align*}
since the $X_i$ are predictable. Then, similar to before, we have that:
\begin{align*}
    \mathbb{E}\Big[ (T_{i}^k)_{+} | \mathscr{F}_{i - 1}\Big] &= \mathbb{E}[||2W_i + 2\Theta^{*}X_i - (\Theta + \Theta^{*})X_i||^k]\mathbb{E}[ ||(\Theta - \Theta^{*})X_i||^k | \mathscr{F}_{i - 1}] \\
    & \leq 2^{2k - 1} \mathbb{E}[||W_i||^k + (KL)^k]\mathbb{E}[||(\Theta - \Theta^{*})X_i||^k | \mathscr{F}_{i - 1}] \\
    & \leq 2^{2k - 1} [\sigma^2 k! Q^{k - 2} + (KL)^k](2KL)^{k-2} \mathbb{E}[||(\Theta - \Theta^{*})X_i||^2 | \mathscr{F}_{i - 1}] \\
    & \leq \frac{[\sigma^2 k! Q^{k - 2} + (KL)^k](8KL)^{k-2}v_i}{\sigma^2 + 4K^2L^2}  \\
    & \leq \frac{k! (8KL(Q \vee KL))^{k-2} v_i}{2}.
\end{align*}
Letting $w = 8KL(Q \vee KL)$, we have that:
\begin{align}
    \mathbb{E}[e^{\xi T_i} | \mathscr{F}_{i - 1}] & \leq 1 + \xi\mathbb{E}[ T_i | \mathscr{F}_{i - 1}] + \sum_{k = 2}^{\infty} \frac{\mathbb{E}[ \xi^k T^k_i | \mathscr{F}_{i - 1}]}{k!} \nonumber \\
    & \leq 1 - \xi  ||(\Theta - \Theta^{*})X_i||^2 + \frac{C||(\Theta - \Theta^{*})X_i||^2 \xi^2}{2} \sum_{k = 0}^{\infty} (\xi w)^k \nonumber \\
    & = 1 - \xi  ||(\Theta - \Theta^{*})X_i||^2 + \frac{C||(\Theta - \Theta^{*})X_i||^2 \xi^2}{2(1 - \xi w)} \nonumber \\
    & \leq \text{exp}\Big(||(\Theta - \Theta^{*})X_i||^2\Big(-\xi + \frac{\xi^2 C}{2(1 - \xi w)}\Big)\Big), \label{eq: Exponential Bound}
\end{align}
where $C = \frac{v_i}{||(\Theta - \Theta^{*})X_i||^2}$. Fix $\xi > 0$ and $\Theta$, and define the process:
\begin{equation*}
    M^{\Theta}_t = \text{exp}\Big(\sum_{i = 1}^t \xi T_i - ||(\Theta - \Theta^{*})X_i||^2\Big(-\xi + \frac{\xi^2 C}{2(1 - \xi w)}\Big) \Big),
\end{equation*}
then, we have from \eqref{eq: Exponential Bound} that $M^{\Theta}_t$ is a supermartingale. Substituting $\xi = \frac{\lambda}{t}$, abbreviating $R(\theta) = \frac{1}{t}\sum_{i = 1}^t ||(\Theta - \Theta^{*})X_i||^2$, observing $\frac{1}{t}\sum_{i = 1}^t T_i = r(\Theta^{*}) - r(\Theta)$, and writing $\Theta = Z(\bar{\Theta})$ for $\bar{\Theta} \in \text{supp}(\mu)$, we have that:
\begin{align}
    \int M^{Z(\bar{\Theta})}_t d\mu(\bar{\Theta}) & = \int \text{exp}\Big(\lambda \Big(r(\Theta^{*}) - r(Z(\bar{\Theta})) + \Big(1 - \frac{C\lambda}{2(t - \lambda w)}\Big) (R(Z(\bar{\Theta})) - R(\Theta^{*})) \Big) - \log \frac{d\rho^{\lambda}_t}{d\mu} \Big)d\rho^{\lambda}_t(\bar{\Theta}) \nonumber \\
    & \geq \text{exp}\Big(\mathbb{E}_{\rho_t^{\lambda}}\Big[\lambda \Big(r(\Theta^{*}) - r(Z(\bar{\Theta})) + \Big(1 - \frac{C\lambda}{2(t - \lambda w)}\Big) (R(Z(\bar{\Theta})) - R(\Theta^{*}))\Big)\Big] - \mathcal{D}_{\text{KL}}(\rho_t^{\lambda}||\mu) \Big), \label{eq: Bounding Martingales}
\end{align}
where we have applied Jensen's inequality in the last step. Now, consider the stopping time $\tau$, defined such that:
\begin{equation*}
    \tau = \inf\Big\{t > 0: \mathbb{E}_{\rho_t^{\lambda}}\Big[\lambda \Big(r(\Theta^{*}) - r(Z(\bar{\Theta})) + \Big(1 - \frac{C\lambda}{2(t - \lambda w)}\Big) (R(Z(\bar{\Theta})) - R(\Theta^{*}))\Big)\Big] - \mathcal{D}_{\text{KL}}(\rho_t^{\lambda}||\mu) - \log \frac{2}{\epsilon} \geq 0 \Big\},
\end{equation*}
where the outer event is defined with respect to the noise $\{\xi_i\}_{i = 1}^{t-1}$, and $r(Z(\bar{\Theta})), R(Z(\bar{\Theta}))$ are predictable in $t$ by definition. Observe, that $\tilde{M}_t = \mathbb{E}_{\mu(\bar{\Theta})}[M^{Z(\bar{\Theta})}_t | \mathscr{F}_{\infty}]$ is also a supermartingale, and hence the stopped process $\tilde{M}_{\tau \wedge t}$ is also supermartingale with $\mathbb{E}[\tilde{M}_{\tau}] \leq 1$ (by Fatou's lemma). We clarify that the former expectation is taken with respect to $\bar{\Theta}$ while \textit{conditioning} on the filtration $\{\mathscr{F}_t\}_{t = 1}^{\infty}$, while the latter expectation is taken with respect to the randomness of the noise $\{\xi_t\}_{t = 1}^{\infty}$ (which \textit{generates} the filtration $\{\mathscr{F}_t\}_{t = 1}^{\infty}$). Define the second process:
\begin{equation*}
    N_t = \text{exp}\Big(\mathbb{E}_{\rho_t^{\lambda}}\Big[ \lambda \Big(r(\Theta^{*}) - r(Z(\bar{\Theta})) + \Big(1 - \frac{C\lambda}{2(t - \lambda w)}\Big) (R(Z(\bar{\Theta})) - R(\Theta^{*}))\Big)\Big] - \mathcal{D}_{\text{KL}}(\rho_t^{\lambda}||\mu) - \log \frac{2}{\epsilon}\Big),
\end{equation*}
used in the definition of $\tau$ above. Observe, from \eqref{eq: Bounding Martingales} that $N_{\tau \wedge t} \leq \frac{\epsilon}{2} \cdot \tilde{M}_{\tau \wedge t}$, and therefore $\mathbb{E}[N_{\tau}] \leq \frac{\epsilon}{2}$. Therefore, by a standard argument, we have that: 
\begin{align*}
    P(\tau < \infty) & \leq P\Big(\log N_{\tau} \geq 0 \Big) \\
    & \leq \mathbb{E}[N_{\tau}] \\
    & \leq \frac{\epsilon}{2},
\end{align*}
and therefore, we have:
\begin{small}
\begin{align*}
    & P\Big(\Big\{\int \Big(r(\Theta^{*}) - r(Z(\bar{\Theta})) + \Big(1 - \frac{C\lambda}{2(t -\lambda w)}\Big) (R(Z(\bar{\Theta})) - R(\Theta^{*}))\Big)d\rho_t^{\lambda}(\Theta) - \frac{1}{\lambda}\Big(\mathcal{D}_{\text{KL}}(\rho_t^{\lambda}||\mu) + \log \frac{2}{\epsilon}\Big) \leq 0: \forall t \geq 1 \Big\}\Big) \\
    & \hspace{450pt} \geq 1 - \frac{\epsilon}{2},
\end{align*}
\end{small}
We clarify that the outer probability is taken with respect to the noise/policy while the inner expectation is taken with respect to the prior. Now applying Jensen's inequality again we have that:
\begin{small}
\begin{align*}
    P\Big(\Big\{\Big(1 - \frac{C\lambda}{2(t -\lambda w)}\Big) & (R(\hat{\Theta}_t) - R(\Theta^{*})) \\
    & \hspace{20pt} \leq \int  \Big(r(Z(\bar{\Theta})) - r(\Theta^{*})\Big)d\rho_t^{\lambda}(\bar{\Theta}) + \frac{1}{\lambda}\Big(\mathcal{D}_{\text{KL}}(\rho_t^{\lambda}||\mu) + \log \frac{2}{\epsilon}\Big) : \forall t \geq 1 \Big\}\Big) \\
    & \hspace{310pt} \geq 1 - \frac{\epsilon}{2}, 
\end{align*}
\end{small}
From the definition of $\rho^{\lambda}_t$ and the Donsker-Varadhan variational formula, we have:
\begin{small}
\begin{align*}
    P\Big(\Big\{\Big(1 - \frac{C\lambda}{2(t -\lambda w)}\Big) & (R(\hat{\Theta}_t) - R(\Theta^{*})) \\
    & \hspace{0pt} \leq \inf_{\rho \in \mathcal{M}(\mu)} \int  \Big(r(Z(\bar{\Theta})) - r(\Theta^{*})\Big)d\rho(\bar{\Theta}) + \frac{1}{\lambda}\Big(\mathcal{D}_{\text{KL}}(\rho||\mu) + \log \frac{2}{\epsilon}\Big) : \forall t \geq 1 \Big\}\Big) \\
    & \hspace{310pt} \geq 1 - \frac{\epsilon}{2},
\end{align*}
\end{small}
where $\mathcal{M}(\mu)$ denotes the set of measures absolutely continuous with respect to $\mu$. Let $\beta = 1 - \frac{C\lambda}{2(t -\lambda w)}$. Like in the proof of Theorem 3.1 in \cite{alquier2011pac}, we wish to bound $\int  \left(r(Z(\bar{\Theta})) - r(\Theta^{*})\right)d\rho(\bar{\Theta})$ by an expression involving the true risk $R$. This is achieved by simply repeating the previous steps to obtain a Bernstein bound for $-T_i$, integrating against $\mu$, and then again performing a change of measure, this time to an arbitrary $\rho \in \mathcal{M}(\mu)$. We obtain:
\begin{small}
\begin{align*}
    P\Big(\Big\{\Big(1 + \frac{C\lambda}{2(t -\lambda w)}\Big) & \Big(\int R(Z(\bar{\Theta})) d\rho(\bar{\Theta}) - R(\Theta^{*})\Big) \\ 
    & \hspace{50pt} \geq \int  \left(r(Z(\bar{\Theta})) - r(\Theta^{*})\right)d\rho(\bar{\Theta}) - \frac{1}{\lambda}\Big(\mathcal{D}_{\text{KL}}(\rho||\mu) + \log \frac{2}{\epsilon}\Big) : \forall t \geq 1 \Big\}\Big) \\
    & \hspace{330pt} \geq 1 - \frac{\epsilon}{2},
\end{align*}
\end{small}
Putting these together and letting $\gamma = 1 + \frac{C\lambda}{2(t -\lambda w)}$, we have, for all $t \geq 1$:
\begin{equation*}
    \beta (R(\hat{\Theta}_t) - R(\Theta^{*})) \leq \inf_{\rho \in \mathcal{M}(\mu)} \gamma\Big(\int R(Z(\bar{\Theta})) d\rho(\bar{\Theta}) - R(\Theta^{*})\Big) + \frac{2}{\lambda}\Big(\mathcal{D}_{\text{KL}}(\rho||\mu) + \log \frac{2}{\epsilon} \Big),
\end{equation*}
with probability $1 - \epsilon$. Now, like in the proof of Theorem 3.1 in \cite{alquier2011pac}, if we take $\lambda = \frac{t}{2\mathcal{C}_1}$ where $\mathcal{C}_1 = C \vee w$, then we have that $\beta \geq \frac{1}{2}$ and $\gamma \leq \frac{3}{2}$, and hence we obtain:
\begin{equation*}
    R(\hat{\Theta}_t) - R(\Theta^{*}) \leq \inf_{\rho \in \mathcal{M}(\mu)} 3\Big(\int R(Z(\bar{\Theta})) d\rho(\bar{\Theta}) - R(\Theta^{*})\Big) + \frac{8\mathcal{C}_1}{t}\Big(\mathcal{D}_{\text{KL}}(\rho||\mu) + \log \frac{2}{\epsilon} \Big),
\end{equation*}
with probability $1 - \epsilon$. The result follows from noting that $R(\Theta^{*}) = 0$ by definition. 
\end{proof}
\begin{proof}{Proof of Lemma \ref{Adapted Case}.}
The key idea is to choose an appropriate measure $\rho \in \mathcal{M}(\mu)$ to estimate the RHS of \eqref{Confidence Set Formula}. We choose $\rho = \frac{\mu_{J^{*}}(\Theta) \mathbbm{1}_{\{||Z(\Theta) - \Theta^{*}|| < \delta\}}}{\mu_J(\{||Z(\Theta) - \Theta^{*}|| < \delta\})}$. Then, we have that:
\begin{align*}
    \frac{d\rho}{d\mu}(\Theta) &= \frac{\mu_{J^{*}}(\Theta)\mathbbm{1}_{\{||Z(\Theta) - \Theta^{*}|| < \delta\}}}{\sum_{J} \pi_{J}\mu_{J}(\Theta)}  \cdot \frac{1}{ \mu_{J^{*}}(\{\Theta: ||Z(\Theta) - \Theta^{*}|| < \delta\})}\\
    & \leq \Big(\pi_{J^{*}}\mu_{J^{*}}(\{\Theta: ||Z(\Theta) - \Theta^{*}|| < \delta\})\Big)^{-1} \\
    & < \infty.
\end{align*}
And hence we have that $\rho \in \mathcal{M}(\mu)$. Observe that since the support of $\rho$ is contained in the event $||Z(\Theta) - \Theta^{*}|| \leq \delta$ by definition, we have:
\begin{equation}
\label{Empirical Risk 0}
    \int R(Z(\Theta)) d\rho(\Theta) \leq L^2\delta^2.
\end{equation}
Moreover, we observe that, by the definition of the transform $Z(\Theta) = \frac{K\Theta}{||\Theta||}$:
\begin{align*}
    \mu_{J^{*}}(\{\Theta: ||Z(\Theta) - \Theta^{*}|| < \delta\}) & \geq  \mu_{J^{*}}(\{\Theta: ||\Theta - \Theta^{*}|| < \delta\}) \\
    & \geq \mu_{J^{*}}(\{\Theta: ||\Theta - \Theta^{*}||_{\text{Fro}} < \delta\}).
\end{align*}
Moreover, if $\delta < 1$, then we observe that $||\Theta - \Theta^{*}||_{\text{Fro}} < \delta$ implies $||\Theta||^2_{\text{Fro}} \leq \delta^2 + ||\Theta^{*}||^2_{\text{Fro}} < K^2N + 1$.  Hence, by the definition of $\mu_{J^{*}}$, we have:
\begin{equation*}
    \mu_{J^{*}}(\{\Theta: ||\Theta - \Theta^{*}||_{\text{Fro}} < \delta\}) = \frac{m_{J^{*}}(\delta \hspace{1mm} \mathbb{B}_{|J^{*}|})}{m_{J^{*}}(\sqrt{K^2 N + 1} \hspace{1mm} \mathbb{B}_{|J^{*}|})},
\end{equation*}
where again $m_{J^{*}}$ denotes the Lebesgue measure on $\mathbb{R}^{|J^{*}|}$ and $\mathbb{B}_{|J^{*}|}$ the unit ball on this space. Hence, we have that:
\begin{align*}
    \text{D}_{\text{KL}}\Big(\rho || m \Big) & \leq -\log {\mu_{J^{*}}(\{\Theta: ||\Theta - \Theta^{*}||_{\text{Fro}} < \delta\})} - \log \pi_{J^{*}} \\
    & = -\log{\frac{m_{J^{*}}(\delta \hspace{1mm} \mathbb{B}_{|J^{*}|})}{m_{J^{*}}(\sqrt{K^2 N + 1} \hspace{1mm} \mathbb{B}_{|J^{*}|})}} - \log \pi_{J^{*}} \\
    & = \log{\Big(\frac{\sqrt{K^2 N + 1}}{\delta}\Big)^{|J^{*}|}} - \log \pi_{J^{*}} \\
    & = |J^{*}|\log{\Big(\frac{\sqrt{K^2 N + 1}}{\delta}\Big)} + \log{\sum_{i=1}^{N^2} \alpha^i} - \log \alpha + \log {N^2 \choose |J^{*}|} \\
    & \leq |J^{*}|\log{\Big(\frac{\sqrt{K^2 N + 1}}{\delta}\Big)} - \log{(1 - \alpha)} - \log \alpha + |J^{*}|\log {\frac{eN^2}{|J^{*}|}}.
\end{align*}
Hence, combining the above result with \eqref{Empirical Risk 0} and substituting into \eqref{Confidence Set Formula} we obtain:
\begin{equation*}
\frac{1}{t}\sum_{s=1}^t \|(\hat{\Theta}_t - \Theta^{*}) p_s\|^2 \leq 3L^2\delta^2 + \frac{8\mathcal{C}_1}{t}\Big(|J^{*}|\log{\Big(\frac{eN^2\sqrt{K^2 N + 1}}{|J^{*}|\delta}\Big)} - \log{(1 - \alpha)} - \log \alpha  + \log \frac{2}{\epsilon} \Big).
\end{equation*}
Substituting $\delta = \frac{1}{t}$, we obtain our result. 
\end{proof}
\begin{proof}{Proof of Lemma \ref{Orders Ellipsoid}.}
The key idea is to choose $\rho \in \mathcal{M}(\mu)$ to estimate the RHS of \eqref{Confidence Set Formula}. For any $\delta > 0$, a natural choice for $\mu$ is:
\begin{equation*}
    \rho(\Theta) = \frac{\sum_{J^{*} \subset J} \pi_J \mu_{J}(\Theta)\mathbbm{1}_{\{||Z(\Theta) - \Theta^{*}|| < \delta\}}(\Theta)}{\sum_{J^{*} \subset J} \pi_J \mu_J({\{||Z(\Theta) - \Theta^{*}|| < \delta\}})}.
\end{equation*}
Then, we have:
\begin{align*}
    \frac{d\rho}{d\mu}(\Theta) &= \frac{\sum_{J^{*} \subset J} \mu_{J}(\Theta)\mathbbm{1}_{\{||Z(\Theta) - \Theta^{*}|| < \delta\}}}{\sum_{J \in \mathcal{J}} \pi_J\mu_{J}(\Theta)}  \cdot \frac{1}{ \sum_{J^{*} \subset J} \mu_{J}(\{\Theta: ||Z(\Theta) - \Theta^{*}|| < \delta\})}\\
    & \leq \Big(\sum_{J^{*} \subset J} \mu_{J}(\{\Theta: ||Z(\Theta) - \Theta^{*}|| < \delta\})\Big)^{-1} \\
    & < \infty,
\end{align*}
And hence we have that $\rho \in \mathcal{M}(\mu)$. Observe that since the support of $\rho$ is contained in the event $||Z(\Theta) - \Theta^{*}|| \leq \delta$ by definition, we have:
\begin{equation}
\label{Risk Bound 0}
    \int R(Z(\Theta)) d\rho(\Theta) \leq L^2\delta^2.
\end{equation}
since $R(\Theta) = \frac{1}{t}\sum_{i = 1}^t ||(\Theta - \Theta^{*})p_i||^2$. Moreover, we observe that, by the definition of the transform $Z(\Theta) = \frac{K\Theta}{||\Theta||}$:
\begin{align*}
    \mu_{J}(\{\Theta: ||Z(\Theta) - \Theta^{*}|| < \delta\}) & \geq  \mu_{J}(\{\Theta: ||\Theta - \Theta^{*}|| < \delta\}) \\
    & \geq \mu_{J}(\{\Theta: ||\Theta - \Theta^{*}||_{\text{Fro}} < \delta\}),
\end{align*}
and hence:
\begin{equation*}
    \mathcal{D}_{KL}(\rho || \mu) \leq \mathbb{E}_{\rho}\Big[-\log \Big(\sum_{J^{*} \subset J} \pi_J \mu_{J}(\{\Theta: ||\Theta - \Theta^{*}|| < \delta\})\Big)\Big].
\end{equation*}
In order to estimate the small ball probability $\mu_{J}(\{\Theta: ||\Theta - \Theta^{*}||^2_{\text{Fro}} < \delta^2\})$, we first bound it from below by a ``centered'' probability. Since the $\Theta_{ij}$ are independent for all $(i, j) \in \mathcal{J}^{*}$ and $||\Theta - \Theta^{*}||^2_{\text{Fro}} = \sum_{(i, j)} (\Theta_{ij} - \Theta^{*}_{ij})^2$, we may bound the density of each squared deviation $(\Theta_{ij} - \Theta^{*}_{ij})^2$ separately. We first consider the case where $i \neq j$. Since $\Theta_{ij} \sim \sqrt{\Gamma(\frac{1}{N}, 1)}$, it is easily verified that the density $\tilde{f}_{ij}$ of $(\Theta_{ij} - \Theta^{*}_{ij})^2$ is:
\begin{equation*}
    \tilde{f}_{ij}(x) = \frac{1}{2\Gamma\left(\frac{1}{N}\right)}\left[(\sqrt{x} + \Theta^{*}_{ij})^{\frac{2}{N} - 2}\left|\frac{\sqrt{x} + \Theta^{*}_{ij}}{\sqrt{x}}\right| e^{-(\sqrt{x} + \Theta^{*}_{ij})^2} + (\sqrt{x} - \Theta^{*}_{ij})^{\frac{2}{N} - 2}\left|\frac{\sqrt{x} - \Theta^{*}_{ij}}{\sqrt{x}}\right|e^{-(\sqrt{x} - \Theta^{*}_{ij})^2}\right].
\end{equation*}
Observe that on the event $\{\Theta: ||\Theta - \Theta^{*}||^2_{\text{Fro}} < \delta^2\}$, we must have $(\Theta_{ij} - \Theta^{*}_{ij})^2 < \delta^2$. Hence, if $0 < \delta < \frac{1}{2} \min_{(i, j) \in J^{*}} |\Theta^{*}_{ij}|$, then we have that $(|\Theta_{ij} - \Theta^{*}_{ij}| + \Theta^{*}_{ij})^2, (|\Theta_{ij} - \Theta^{*}_{ij}| - \Theta^{*}_{ij})^2 < \frac{9}{4}(\Theta^{*}_{ij})^2 < \frac{1}{2}$ by the assumption that $N > \frac{3C}{\sqrt{2}}$ (and hence $(\Theta^{*}_{ij})^2 \leq \Big(\frac{C}{N}\Big)^2 < \frac{2}{9}$). Thus, since the function $x^{\frac{1}{N} - 1}e^{x}$ is convex on $(0, 0.5)$ and $\Big|\frac{\sqrt{x} \pm \Theta^{*}_{ij}}{\sqrt{x}}\Big| \geq 1$, when $\sqrt{x} < \delta < \frac{1}{2} \min_{(i, j) \in J^{*}} |\Theta^{*}_{ij}|$, we  have that
\begin{align*}
    \tilde{f}_{ij}(x) & = \frac{1}{2\Gamma\left(\frac{1}{N}\right)}\left[(\sqrt{x} + \Theta^{*}_{ij})^{\frac{2}{N} - 2}\left|\frac{\sqrt{x} + \Theta^{*}_{ij}}{\sqrt{x}}\right| e^{-(\sqrt{x} + \Theta^{*}_{ij})^2} + (\sqrt{x} - \Theta^{*}_{ij})^{\frac{2}{N} - 2}\left|\frac{\sqrt{x} - \Theta^{*}_{ij}}{\sqrt{x}}\right|e^{-(\sqrt{x} - \Theta^{*}_{ij})^2}\right] \\
    & \geq \frac{1}{2\Gamma\left(\frac{1}{N}\right)}\left[(\sqrt{x} + \Theta^{*}_{ij})^{\frac{2}{N} - 2} e^{-(\sqrt{x} + \Theta^{*}_{ij})^2} + (\sqrt{x} - \Theta^{*}_{ij})^{\frac{2}{N} - 2}e^{-(\sqrt{x} - \Theta^{*}_{ij})^2}\right] \\
    & \geq \frac{1}{\Gamma\left(\frac{1}{N}\right)}x^{\frac{1}{N} - 1}e^{-x - (\Theta^{*}_{ij})^2}
\end{align*}
on the event $\{\Theta: ||\Theta - \Theta^{*}||^2_{\text{Fro}} < \delta^2\}$ (where the last line follow from Jensen's inequality). Similarly, when $i = j$, we have that $\Theta_{ii} \sim |N(0, 1)|$, and hence $(\Theta_{ii} - \Theta^{*}_{ii})^2$ has a noncentral $\chi^2$ distribution with noncentrality parameter $\Theta^{*}_{ii}$. Hence, its density is given by:
\begin{align*}
    \tilde{f}_{ii}(x) & = \frac{x^{-\frac{1}{2}}}{2\sqrt{2\pi}}(e^{-(\sqrt{x} + \Theta^{*}_{ii})^2} + e^{-(\sqrt{x} - \Theta^{*}_{ii})^2}) \\
    & \geq \frac{1}{\sqrt{2\pi}}e^{-x - (\Theta^{*}_{ij})^2}
\end{align*}
on the event $\{||\Theta - \Theta^{*}||^2_{\text{Fro}} < \delta^2\}$. Here the last line again follows by Jensen's inequality and the fact that $\sqrt{x} < \delta < \frac{1}{2}$ since $\min_{(i,j) \in J^{*}} |\Theta_{ij}| \leq \frac{C}{N} < 1$ as $N > \frac{3C}{\sqrt{2}}$ by assumption. Putting this all together, we have:
\begin{equation*}
    \mu_{J}(||\Theta - \Theta^{*}||^2_{\text{Fro}} < \delta) \geq  e^{-||\Theta^{*}||^2_{|\text{Fro}}}\mu_{J}(||\Theta||^2_{\text{Fro}} < \delta).
\end{equation*}
We  may now estimate the small ball probability $\mu_{J}(||\Theta||^2_{\text{Fro}} < \delta)$ using the exponential Tauberian theorem (see Theorem 3.5 in \cite{li2001gaussian} or Theorem 4.2.19 in \cite{bingham1989regular}). Hence, we first study the asymptotic behavior of the cumulant generating function of $||\Theta||^2_{\text{Fro}}$. Namely, for $\lambda > 0$:
\begin{align*}
    \mathbb{E}\Big[\text{exp}\Big(-\lambda||\Theta||^2_{\text{Fro}}\Big)\Big] &= \Big(\prod_{(i, j) \in J, i \neq j} \mathbb{E}[\text{exp}(-\lambda \Theta^2_{ij})]\Big) \Big(\prod_{(i, i) \in J} \mathbb{E}\Big[\text{exp}(-\lambda \Theta^2_{ii})\Big]\Big) \\
    &  = \prod_{(i, j) \in J, i \neq j} (1 + \lambda)^{-\frac{1}{N}} \cdot \prod_{(i, i) \in J} (1 + 2\lambda)^{-1}.
\end{align*}
Letting $J_{\text{OD}} = |\{(i, j) \in J| i \neq j\}|$ and $J_{\text{D}} = |\{(i, i) \in J\}|$, we have that:
\begin{align*}
    \log \mathbb{E}\Big[\text{exp}\Big(-\lambda||\Theta||^2_{\text{Fro}}\Big)\Big] & = -\frac{1}{N}\sum_{(i, j) \in J; i \neq j} \log{(1 + \lambda)} -  \sum_{(i, i) \in J} \log{(1 + 2\lambda)} \\
    & = -\frac{J_{\text{OD}}}{N} \cdot \log{(1 + \lambda)} - J_{\text{D}}\log{(1 + 2\lambda)} \\
    & \asymp \Big(\frac{J_{\text{OD}}}{N} + J_{\text{D}}\Big) \log{\lambda}.
\end{align*}
Hence, by Theorem 3.5 in \cite{li2001gaussian}, we have that:
\begin{equation*}
    \log P(||\Theta||^2_{\text{Fro}} < \delta) \asymp \Big(\frac{J_{\text{OD}}}{N} + J_{\text{D}}\Big) \log \delta \geq (2N - 1) \log \delta,
\end{equation*}
since $J_{\text{OD}} \leq N^2 - N$ and $\delta < 1$. Finally, we note that:
\begin{align*}
    -\log{\sum_{J^{*} \subset J} \pi_{J}} & \leq -\log {\pi_{[N] \times [N]}} \\
    & \leq \log{\sum_{i = 1}^N i^{-2}} + 4 \log N \\
    & \leq  \log{\frac{\pi^2}{6}} + 4 \log N.
\end{align*}
since $\pi_{[N] \times [N]} = \frac{N^{-4}}{\sum_{i = 1}^N i^{-2}}$ by definition \eqref{Specified Sparsity Mixing}. Putting this all together, we have:
\begin{align*}
    \mathcal{D}_{KL}(\rho || \mu) & \leq ||\Theta^{*}||^2_{\text{Fro}} + \log{\frac{\pi^2}{6}} + 4 \log N - (2N - 1) \log \delta  \\
    & \leq K^2 N + \log{\frac{\pi^2}{6}} + 4 \log N - (2N - 1) \log \delta \\
    & \leq K^2 N + \log{\frac{\pi^2}{6}} + 4 \log N + (2N - 1)\log t - (2N - 1) \log{\min_{(i, j) \in J^{*}} |\Theta^{*}_{ij}|},
\end{align*}
where in the last step we have substituted $\delta = \frac{\min_{(i, j) \in J^{*}} |\Theta^{*}_{ij}|}{t}$. Now, like before plugging \eqref{Risk Bound 0} and the above display into \eqref{Confidence Set Formula} we obtain our desired result. 
\end{proof}
\begin{proof}{Proof of Lemma \ref{Spectral Scaling Confidence, Exponential}.}
The key idea again is to choose $\rho \in \mathcal{M}(\mu)$ to estimate the RHS of \eqref{Confidence Set Formula}. Here, we choose 
\begin{equation*}
    \rho(\Theta) = \frac{\int_{\gamma \geq 1} \pi(\gamma)\mu_{\gamma}(\Theta)\mathbbm{1}_{\{||Z(\Theta) - \Theta^{*}|| < \delta\}}d\gamma}{\int_{\gamma \geq 1} \mu_{\gamma}(\{\Theta: ||Z(\Theta) - \Theta^{*}|| < \delta\})\pi(\gamma)d\gamma}.
\end{equation*}
Then, we have:
\begin{align*}
    \frac{d\rho}{d\mu}(\Theta) &= \frac{\int_{\gamma \geq 1} \pi(\gamma)\mu_{\gamma}(\Theta)\mathbbm{1}_{\{||Z(\Theta) - \Theta^{*}|| < \delta\}}d\gamma}{\int_{\gamma > 0} \pi(\gamma)\mu_{\gamma}(\Theta)d\gamma}  \cdot \frac{1}{\int_{\gamma \geq 1} \mu_{\gamma}(\{\Theta: ||Z(\Theta) - \Theta^{*}|| < \delta\})\pi(\gamma)d\gamma}\\
    & \leq \Big(\int_{\gamma \geq 1} \mu_{\gamma}(\{\Theta: ||Z(\Theta) - \Theta^{*}|| < \delta\})\pi(\gamma)d\gamma\Big)^{-1} \\
    & < \infty,
\end{align*}
and hence we have that $\rho \in \mathcal{M}(\mu)$. Observe that since the support of $\rho$ is contained in the event $||Z(\Theta) - \Theta^{*}|| \leq \delta$ by definition, we have:
\begin{equation}
\label{Risk Bound 1}
    \int R(\Theta) d\rho(\Theta) \leq L^2\delta^2,
\end{equation}
since $R(\Theta) = \frac{1}{t}\sum_{i = 1}^t ||(\Theta - \Theta^{*})p_i||^2$. Moreover, we observe that, by the definition of the transform $Z(\Theta) = \frac{K\Theta}{||\Theta||}$:
\begin{align*}
    \mu_{\gamma}(\{\Theta: ||Z(\Theta) - \Theta^{*}|| < \delta\}) & \geq  \mu_{\gamma}(\{\Theta: ||\Theta - \Theta^{*}|| < \delta\}) \\
    & \geq \mu_{\gamma}\Big(\Big\{\kappa: ||\kappa - \kappa^{*}||_{\infty} < \frac{\delta}{N}\Big\}\Big),
\end{align*}
where the last line follows from the fact that for any $N \times N$ matrix $A$, $||A|| \leq N \max_{1 \leq i, j \leq N} A_{ij}$. Hence, we have that:
\begin{equation}
\label{KL Formula}
    \mathcal{D}_{KL}(\rho || \mu) \leq \mathbb{E}_{\rho}\Big[-\log \Big(\int_{\gamma \geq 1} \mu_{\gamma}\Big(\Big\{\kappa: ||\kappa - \kappa^{*}||_{\infty} < \frac{\delta}{N}\Big\}\Big) d\gamma\Big)\Big].
\end{equation}
In order to approximate the latter small ball probabilities, we first note that $\mathcal{H}^{\alpha} = [\mathcal{H}^{\beta}, \mathcal{H}]_{\frac{\alpha - \beta}{1 - \beta}, 2}$ by the reiteration property of real interpolation (Theorem V.2.4 in \cite{bennett1988interpolation}). Moreover, by Proposition V.1.10 in \cite{bennett1988interpolation}, we have that $\mathcal{H}^{\alpha} = [\mathcal{H}^{\beta}, \mathcal{H}]_{\frac{\alpha - \beta}{1 - \beta}, 2} \subset [\mathcal{H}^{\beta}, \mathcal{H}]_{\frac{\alpha - \beta}{1 - \beta}, \infty}$. For the first step, we follow an argument similar to that of Theorem 6 in \cite{suzuki2012pac}: since $\kappa^{*} \in \mathcal{H}^{\alpha}$, we have by definition, that for each $t > 0$, there exists a $g_t \in \mathcal{H}$ such that $||\kappa^{*}||_{\frac{\alpha - \beta}{1 - \beta}, \infty} \leq t^{\frac{\beta - \alpha}{1 - \beta}}||\kappa^{*} - g_t||_{\mathcal{H}^{\beta}} + t^{\frac{1 - \alpha}{1 - \beta}}||g_t||_{\mathcal{H}} \leq 2||\kappa^{*}||_{\frac{\alpha - \beta}{1 - \beta}, \infty} < \infty$. Hence, since $t^{\frac{\beta - \alpha}{1 - \beta}} \to \infty$ as $t \to 0$, we have that for every $\delta > 0$, there exists a $g_{\delta} \in \mathcal{H}$ such that $||g_{\delta} - \kappa^{*}||_{\mathcal{H}^{\beta}} < \delta$. Moreover, after noting that $||g_t -  \kappa^{*}||_{\mathcal{H}^{\beta}} \leq 2t^{\frac{\alpha - \beta}{1 - \beta}}||\kappa^{*}||_{\frac{\alpha - \beta}{1 - \beta}, \infty}$ and $||g_t||_{\mathcal{H}} \leq 2t^{\frac{\alpha - 1}{1 - \beta}}||\kappa^{*}||_{\frac{\alpha - \beta}{1 - \beta}, \infty}$, and some algebra, we obtain:
\begin{equation}
\label{Approximation Norm}
    ||g_{\delta}||_{\mathcal{H}} \leq 2^{\frac{1 - \beta}{\alpha - \beta}}||\kappa^{*}||_{\frac{\alpha - \beta}{1 - \beta}, \infty}^{\frac{1 - \beta}{\alpha - \beta}}||g_{\delta} -  \kappa^{*}||_{\mathcal{H}^{\beta}}^{\frac{\alpha - 1}{\alpha - \beta}} \leq 2^{\frac{1 - \beta}{\alpha - \beta}}||\kappa^{*}||_{\frac{\alpha - \beta}{1 - \beta}, \infty}^{\frac{1 - \beta}{\alpha - \beta}}\delta^{\frac{\alpha - 1}{\alpha - \beta}}.
\end{equation}
Since $\mathcal{H}_K^{\beta} \hookrightarrow L_{\infty}$ for some $\beta < \alpha$ by assumption, we have that there exists a $M > 0$ such that:
\begin{align*}
    \mu_{\gamma}(\{\Theta: ||\kappa - \kappa^{*}||_{\infty} < \delta N^{-1}\}) & \geq  \mu_{\gamma}\Big(\Big\{\Theta: ||\kappa - \kappa^{*}||_{\mathcal{H}^{\beta}} < \frac{\delta}{MN}\Big\}\Big) \\
    & \geq \mu_{\gamma}\Big(\Big\{\Theta: ||\kappa - g_{0.5M^{-1}\delta N^{-1}}||_{\mathcal{H}^{\beta}} < \frac{\delta}{2MN}\Big\}\Big).
\end{align*}
We note that the distances in the previous display are well-defined --- indeed, since the RKHS of $\mu_{\gamma}$ is $\mathcal{H}$ (with scaled reproducing kernel $\gamma \mathcal{K}$) and $\beta < 1$, we have that the embedding $i_{\beta}: \mathcal{H} \to \mathcal{H}^{\beta}$ is Hilbert-Schmidt (this can be easily observed from \eqref{Embedding Operator} since $||i_{\beta}||^2_{\text{HS}} = \sum_{i = 1}^{\infty} \mu_i^{1 - \beta} \asymp \sum_{i = 1}^{\infty} e^{-iq(1 - \beta)} < \infty$. Hence, the sample paths of $\mu_{\gamma}$ lie almost surely in $\mathcal{H}^{\beta}$ (see Theorem 4.1 in \cite{steinwart2019convergence}); in addition $\mathcal{H}^{\alpha} \subset \mathcal{H}^{\beta}$ since $\alpha > \beta$. Thus, we have, by Borell's theorem and the fact that $||\cdot||_{\gamma \mathcal{K}} = \gamma^{-1}||\cdot||_{\mathcal{K}}$, that the latter small probability is bounded below by:
\begin{equation}
\label{Borell's theorem}
    \mu_{\gamma}\Big(\Big\{\Theta: ||\kappa - g_{0.5M^{-1}\delta N^{-1}}||_{\mathcal{H}^{\beta}} < \frac{\delta}{2MN}\Big\}\Big) \geq \text{exp}\Big(-\gamma^{-1}||g_{0.5M^{-1}\delta N^{-1}}||^2_{\mathcal{K}}\Big)\mu_{\gamma}\Big(\Big\{\kappa: ||\kappa||_{\mathcal{H}^{\beta}} < \frac{\delta}{2MN}\Big\}\Big).
\end{equation}
Now, writing the Karheunen-Loeve expansion of $\kappa$ as $\kappa = \sum_{i = 1}^{\infty} Z_i\mu_i^{\frac{1}{2}}e_i = \sum_{i = 1}^{\infty} (Z_i\mu_i^{\frac{1 - \beta}{2}})\mu_i^{\frac{\beta}{2}}e_i$ (where $Z_i \sim \mathcal{N}(0, 1)$ i.i.d), observing that $||\kappa^{*}||^2_{\mathcal{H}^{\beta}} = \sum_{i} Z^2_i \mu_i^{(1 - \beta)} \asymp \sum_{i} Z^2_i e^{-iq(1 - \beta)}$, and applying a slight modification of Theorem 4.6 in \cite{dunker1998small} we have that:
\begin{small}
\begin{align*}
    -\log \mu_{\gamma}\Big(\Big\{\kappa: ||\kappa||_{\mathcal{H}^{\beta}} < \frac{\delta}{2MN}\Big\}\Big) & \asymp \frac{\pi^2}{12} + \frac{1}{4q(1 - \beta)}\Big(\log  \Big(\frac{4M^2N^2}{\delta^2} \log \frac{4M^2N^2}{\delta^2}\Big)\Big)^2 - \frac{1}{4}\log \frac{4M^2N^{2}}{\delta^2} + \frac{3}{4} \log \log  \frac{4M^2N^{2}}{\delta^2} \\
    & \asymp \frac{1}{q(1 - \beta)}\log^2  \Big(\frac{2MN}{\delta}\Big).
\end{align*}
\end{small}
as $\delta \to 0$ for some absolute constant $C$. Hence, plugging in $\delta = \frac{1}{\sqrt{t}}$ in the above display, we obtain, via \eqref{KL Formula}, \eqref{Approximation Norm}, \eqref{Borell's theorem}, and $\gamma \geq 1$ that:
\begin{equation*}
    \mathcal{D}_{KL}(\rho || \mu) \leq 4^{\frac{1 - \beta}{\alpha - \beta}}||\kappa^{*}||_{\frac{\alpha - \beta}{1 - \beta}, \infty}^{\frac{2(1 - \beta)}{\alpha - \beta}}(4M^2N^2t)^{\frac{1 - \alpha}{\alpha - \beta}} + \mathcal{C}_{\beta, q}\log^2  \Big(2MN\sqrt{t}\Big) - \log{(\pi([1, \infty)))},
\end{equation*}
where $\mathcal{C}_{\beta, q}$ is a constant depending only on $\beta$ and $q$. Combining this expression with \eqref{Risk Bound 1} and substituting back into \eqref{Confidence Set Formula}, we obtain our result.
\end{proof}
\begin{proof}{Proof of Lemma \ref{Spectral Confidence, Exponential}.}
Like before the key to our argument is to construct an an appropriate measure $\rho \in \mathcal{M}(\mu)$ to estimate the RHS of \eqref{Confidence Set Formula}. Let $\iota \in (0, \alpha - \beta)$. Then we set:
\begin{equation*}
    \rho(\Theta) = \frac{\int_{\beta + \iota}^{\alpha} \pi(\gamma)\mu_{\gamma}(\Theta)\mathbbm{1}_{\{||Z(\Theta) - \Theta^{*}||_{2} < \delta\}}d\gamma}{\int_{\beta + \iota}^{\alpha} \mu_{\gamma}(\{\Theta: ||Z(\Theta) - \Theta^{*}||_{2} < \delta\})\pi(\gamma)d\gamma},
\end{equation*}
where we recall that $\kappa \in \mathcal{H}^{\alpha}$ by assumption. Observe that $\rho \in \mathcal{M}(\mu)$ by a similar logic as in the proof of Lemma \ref{Spectral Scaling Confidence, Exponential} and $\Big|\frac{d\rho}{d\mu}\Big| \leq \Big(\int_{\beta + \iota}^{\alpha} \mu_{\gamma}(\{\Theta: ||Z(\Theta) - \Theta^{*}||_{2} < \delta\})d\gamma\Big)^{-1}$. Like before, by the construction of $\rho$ we have:
\begin{equation*}
    \int R(\Theta) d\rho(\Theta) \leq L^2\delta^2.
\end{equation*}
Moreover, we observe that, by the definition of the transform $Z(\Theta) = \frac{K\Theta}{||\Theta||}$:
\begin{align*}
    \mu_{\gamma}(\{\Theta: ||Z(\Theta) - \Theta^{*}||_{2} < \delta\}) & \geq  \mu_{\gamma}(\{\Theta: ||\Theta - \Theta^{*}||_{2} < \delta\}) \\
    & \geq \mu_{\gamma}\Big(\Big\{\Theta: ||\kappa - \kappa^{*}||_{\infty} < \frac{\delta}{N}\Big\}\Big),
\end{align*}
where the last line follows from the fact that for any $N \times N$ matrix $A$, $||A||_{2} \leq N \max_{(i, j)} |A_{ij}|$. Thus, we have:
\begin{align}
    \mathcal{D}_{\text{KL}}(\rho||\mu) & = \int \log \Big(\frac{d\rho}{d\mu}\Big)d\rho \nonumber \\
    & \leq \mathbb{E}_{\rho}\Big[-\log \Big(\int_{\beta + \iota}^{\alpha} \pi(\gamma)\mu_{\gamma}\Big(\Big\{\kappa: ||\kappa - \kappa^{*}||_{\infty} < \frac{\delta}{N}\Big\}\Big)d\gamma\Big)\Big]. \label{eq: Integrated Small Ball}
\end{align}
We now calculate the small ball probabilities in \eqref{eq: Integrated Small Ball}. We first note that, if $\gamma \in [\beta + \iota, \alpha)$, then ${\sf tr}(T^{\gamma - \beta}_{K}) = \sum_{i = 1} \mu^{\gamma - \beta}_i \asymp \sum_{i = 1}^{\infty} e^{-q(\gamma - \beta)i} \leq \sum_{i = 1}^{\infty} e^{-q\iota i} < \infty$. Thus, the inclusion $J: \mathcal{H}^{\gamma} \to \mathcal{H}^{\beta}$ is Hilbert-Schmidt, and the support of $\mu_{\gamma}$ is contained in $\mathcal{H}^{\beta}$ by Theorem 4.4 in \cite{steinwart2019convergence}. Moreover, since $\mathcal{H}^{\beta} \hookrightarrow L_{\infty}$ by assumption, we have that there exists a $M > 0$ such that:
\begin{equation*}
    \mu_{\gamma}(\{\kappa: ||\kappa - \kappa^{*}||_{\infty} < \delta N^{-1}\}) \geq \mu_{\gamma}(\{\kappa: ||\kappa - \kappa^{*}||_{\mathcal{H}^{\beta}} < \delta(MN)^{-1}\}).
\end{equation*}
We observe that since $\kappa^{*} \in \mathcal{H}^{\alpha}_K \subset \mathcal{H}^{\gamma}_K$ for $\gamma \in [\beta + \iota, \alpha)$ by assumption, it follows that $\kappa^{*}$ belongs to the Cameron-Martin spaces of the measures $\mu_{\gamma}$ (as the spaces $\{\mathcal{H}^{\eta}\}_{\eta \in [0, 1]}$ are decreasing), and therefore the shifted measures $\tilde{\mu}_{\gamma}(\cdot) = \mu_{\gamma}(\cdot - \kappa^{*})$ are absolutely continuous with respect to $\mu_{\gamma}$ by the Cameron-Martin formula (see e.g. Theorem 6.7 in \cite{kukush2020gaussian}). Then, by Borell's theorem, we have:
\begin{align}
    \mu_{\gamma}(\{\kappa: ||\kappa - \kappa^{*}||_{\infty} < \delta N^{-1}\}) & \geq \mu_{\gamma}\Big(\Big\{\kappa: ||\kappa - \kappa^{*}||_{\mathcal{H}^{\beta}} < \frac{\delta}{MN}\Big\}\Big) \nonumber \\
    & \geq \text{exp}\Big(-||\kappa^{*}||^2_{\mathcal{H}^{\gamma}}\Big)\mu_{\gamma}\Big(\Big\{\kappa: ||\kappa||_{\mathcal{H}^{\beta}} < \frac{\delta}{MN}\Big\}\Big). \label{eq: Borell application}
\end{align}
Now, writing the Karheunen-Loeve expansion of $\kappa$ as $\kappa = \sum_{i = 1}^{\infty} Z_i\mu_i^{\frac{\gamma}{2}}e_i = \sum_{i = 1}^{\infty} (Z_i\mu_i^{\frac{\gamma - \beta}{2}})\mu_i^{\frac{\beta}{2}}e_i$ (where $Z_i \sim \mathcal{N}(0, 1)$ i.i.d), observing that $||\kappa^{*}||^2_{\mathcal{H}^{\beta}} = \sum_{i} Z^2_i \mu_i^{(\gamma - \beta)} \asymp \sum_{i} Z^2_i e^{-iq(\gamma - \beta)}$, and again applying a slight modification of Theorem 4.6 in \cite{dunker1998small} we have that:
\begin{small}
\begin{align}
    -\log \mu_{\gamma}\Big(\Big\{\kappa: ||\kappa||_{\mathcal{H}^{\beta}} < \frac{\delta}{MN}\Big\}\Big) & \asymp \frac{\pi^2}{12} + \frac{1}{4q(\gamma - \beta)}\Big(\log  \Big(\frac{M^2N^{2}}{\delta^2} \log \frac{M^2N^2}{\delta^2}\Big)\Big)^2 - \frac{1}{4}\log \frac{M^2N^2}{\delta^2} + \frac{3}{4} \log \log  \frac{M^2N^2}{\delta^2} \nonumber \\
    & \leq \frac{C}{q\iota}\log^2  \Big(\frac{M^2N^2}{\delta^2}\Big), \label{eq: Exp Small Ball}
\end{align}
\end{small}
as $\delta \to 0$ for some absolute constant $C$ (the last line follows from the fact that $\gamma \geq \beta + \iota$). Moreover, we note that, since $\kappa^{*} \in \mathcal{H}^{\alpha}$ by assumption, we may write for $\gamma < \alpha$ that $\kappa^{*} = \sum_{i = 1}^{\infty} c_i \mu_i^{\frac{\alpha}{2}} e_i = \sum_{i = 1}^{\infty} c_i \mu_i^{\frac{\alpha - \gamma}{2}} \mu_i^{\frac{\gamma}{2}} e_i $ for some $(c_i)_{i = 1}^{\infty} \in \ell^2$. Hence, we have in \eqref{eq: Borell application} that:
\begin{align}
    ||\kappa^{*}||^2_{\mathcal{H}^{\gamma}} & = \sum_{i = 1}^{\infty} c^2_i \mu_i^{\alpha - \gamma} e_i \nonumber \\
    & \leq \sum_{i = 1}^{\infty} c^2_i \nonumber \\
    & = ||\kappa^{*}||^2_{\mathcal{H}^{\alpha}}, \label{eq: Norm Dominance}
\end{align}
since $\mu_i \asymp e^{-iq}$ by assumption. Hence, plugging in $\delta = \frac{1}{t}$ in \eqref{eq: Exp Small Ball}, we obtain, via \eqref{Confidence Set Formula}, \eqref{eq: Borell application}, and \eqref{eq: Norm Dominance} that:
\begin{equation*}
    R(\hat{\Theta}_t) - R(\Theta^{*}) \leq  \frac{3L^2}{t^2} + \frac{8\mathcal{C}_1}{t}\Big(\mathcal{C}_{\alpha, \beta}||\kappa^{*}||^2_{\mathcal{H}^{\alpha}} + \mathcal{C}_{\alpha, \beta, q}\log^2  MNt + \log \frac{2}{\epsilon} \Big),
\end{equation*}
where again $\mathcal{C}_{\beta, q, \alpha}$ is a constant the depends only on $\beta, q$, and $\alpha$, and likewise for $\mathcal{C}_{\alpha, \beta}$ (note we may set $\iota$ to any fixed value in $(0, \alpha - \beta)$). 
\end{proof}
\bibliographystyle{informs2014}
\bibliography{ref}
\end{document}